\documentclass[10pt,twocolumn,letterpaper]{article}

\usepackage{wacv}
\usepackage{times}
\usepackage{epsfig}
\usepackage{graphicx}
\usepackage{amsmath}
\usepackage{amssymb}
\usepackage{booktabs}

\usepackage{tikz}
\usepackage{comment}
\usepackage{color}
\usepackage[accsupp]{axessibility}

\usepackage{graphicx}
\usepackage{amsmath}
\usepackage{amssymb}
\usepackage{booktabs}
\usepackage{algpseudocode}
\usepackage{algorithm}
\usepackage{fixmath}
\usepackage{caption}
\usepackage{subcaption}
\usepackage{layouts}

\def\wacvPaperID{763} 

\wacvalgorithmstrack   

\wacvfinalcopy 

\usepackage[pagebackref=true,breaklinks=true,letterpaper=true,colorlinks,bookmarks=false,citecolor=orange]{hyperref}

\begin{document}

\title{TeST: Test-time Self-Training under Distribution Shift}

\author{
Samarth Sinha \\
University of Toronto\\
\and
Peter Gehler\\
AWS Tubingen\\
\and
Francesco Locatello\\
AWS Tubingen\\
\and
Bernt Schiele\\
AWS Tubingen\\
}

\maketitle
\thispagestyle{empty}

\begin{abstract}
Despite their recent success, deep neural networks continue to perform poorly when they encounter
distribution shifts at test time.
Many recently proposed approaches try to counter this by aligning the model to the new
distribution prior to inference. 
With no labels available this requires unsupervised 
objectives to adapt the model on the observed test data. 
In this paper, we propose Test-Time Self-Training (TeST): a technique that takes as input a model trained on some source data and
a novel data distribution at test time, and learns invariant and robust representations using
a student-teacher framework. We find that models adapted using TeST significantly improve over 
baseline test-time adaptation algorithms.
TeST achieves competitive performance to modern domain adaptation algorithms 
\cite{chen2020harmonizing,saito2019strong}, while having access to 5-10x less data at time of adaption.
We thoroughly evaluate a variety of baselines on two tasks: object detection and image segmentation and find that models adapted with TeST.
We find that TeST sets the new state-of-the art for test-time
domain adaptation algorithms. 
\end{abstract}


\section{Introduction}
\label{sec:intro}

Deep learning models have shown excellent promise in computer vision research, where 
models are used to perform core computer vision tasks such as 
image classification \cite{vgg,resnet},
pixel-wise segmentation \cite{fcn,drn,mask_rcnn}, and 
object detection \cite{fast_rcnn,ren2015faster,fpn}.
However, one key limitation still shared with most statistical machine learning models is the limited ability to 
generalize across distribution shift. In distribution shift we face test
data that comes from a different distribution than what the model has been trained on
\cite{gen_1,gen_2}.
More specifically, when there is a covariate shift between the training
distribution $P_{\mathcal{S}}(x)$, and the test distribution $P_{\mathcal{T}}(x)$, 
but the distribution of the classes $P(y|x)$ remains constant, neural networks are unable
to adapt to such novel domains.
This affinity towards the training distribution hinders the ability of the model
to be deployed in real-world settings, as it is impossible to train the model
on all possible data distributions that it may encounter in the real-world.


\begin{figure}[t!]
    \centering
    \includegraphics[width=0.47
    \textwidth]{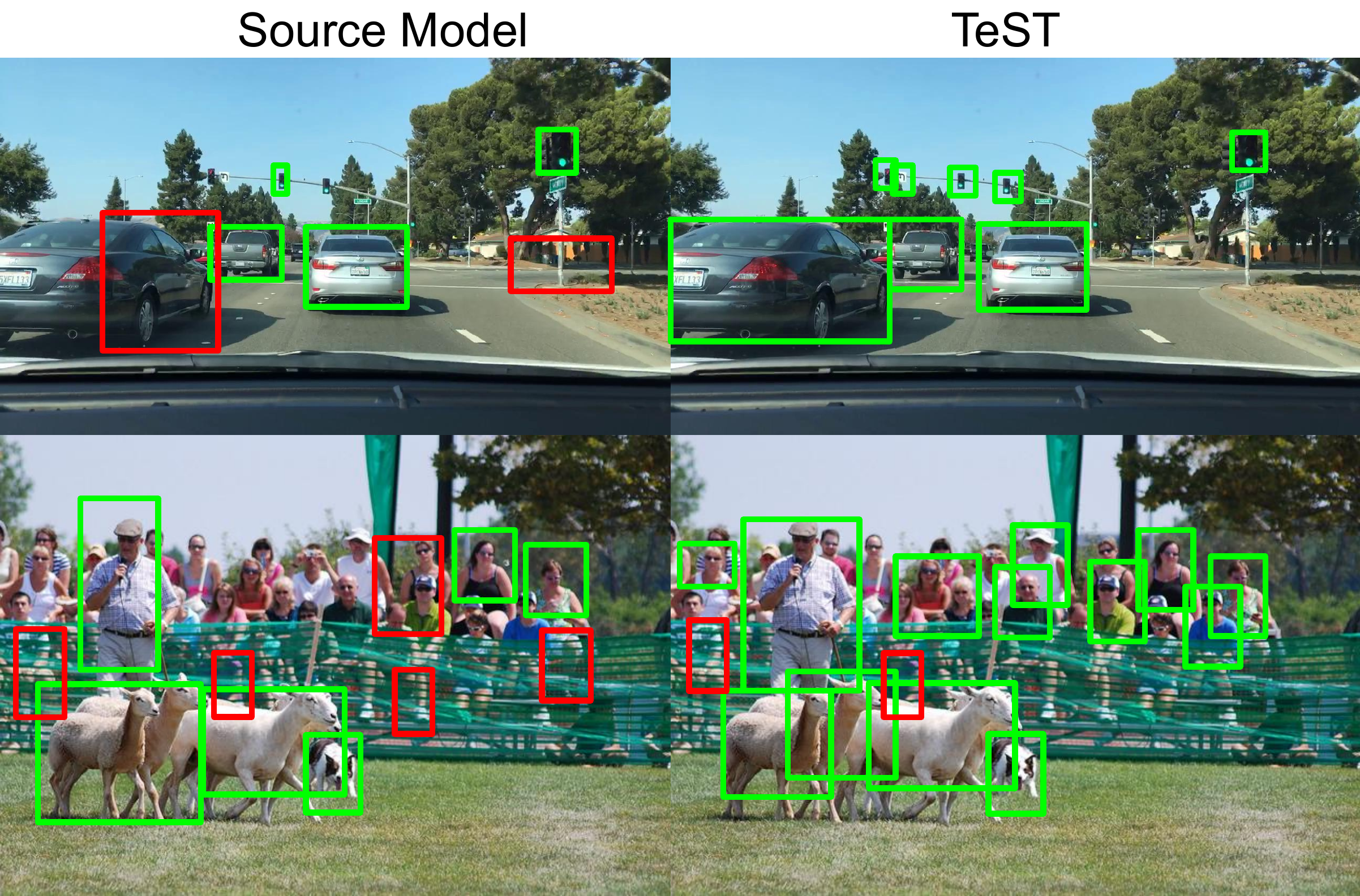}
    \caption{The output of a Faster-RCNN before and after trained on CityScapes (top) and MS-COCO (bottom). 
    Note that the number of false positives ({\color{red} red} rectangles) is reduced while the number of true positives ({\color{green} green} rectangles) is significantly improved. 
    We see that using TeST, we are able to significantly improve the bounding box positions, and also able to 
    capture more objects that are clearly within the scene. For example, in the top image, the TeST model 
    correctly detects traffic signs and traffic lights far from the car, while 
    also improving the bounding box location for the cars. Similarly, the TeST model
    is able to detect more people in the background in the bottom example, as well as able to detect more sheep in the scene.
    }
    \label{fig:intro_teaser}
\end{figure}

Typically, the problem of \textit{domain adaptation} has been addressed by assuming
that the model has access to unlabeled test data during training time
\cite{sup_da_1,sup_da_2,sup_da_3,sup_da_4}.
The downside of this strategy is that the model can only adapt to distributions
for which data is available during training, and collecting data from 
all possible target distributions is often impractical. 
Recently, transductive approaches, also referred to \textit{test-time adaptation}, have garnered significant attention
\cite{ttt,tent,tailoring}.
But such algorithms do not directly solve the problem of learning invariant representations
such that the model can deal with distribution shifts as in the case of domain adaptation.
Towards this, we propose a test-time adaptation method that utilizes consistency 
regularization to learn \emph{invariant representations} \cite{fixmatch,crvae}, 
self-distillation using a student-teacher framework to learn \emph{robust representations} from the
test-data for adaptation \cite{xie2020self}, and entropy minimization to produce 
\emph{confident predictions} on the novel test data \cite{tent}. 
We find that despite its simplicity, our Test-time Self Training (TeST) approach significantly improves the performance of 
the model, as can be seen qualitatively in Figure~\ref{fig:intro_teaser}. We achieve this without \textit{any changes to the training paradigm} (such as having to train
the source model on an additional image rotation loss at test time \cite{ttt})
or access to the source data \cite{saito2019strong,chen2020harmonizing},
and do not require large batch-size \cite{tent}. 
Our only requirements is that the label space is shared by the source and target domain.

\begin{figure*}
    \centering
    \begin{subfigure}{0.49\textwidth}
    \includegraphics[width=\textwidth]{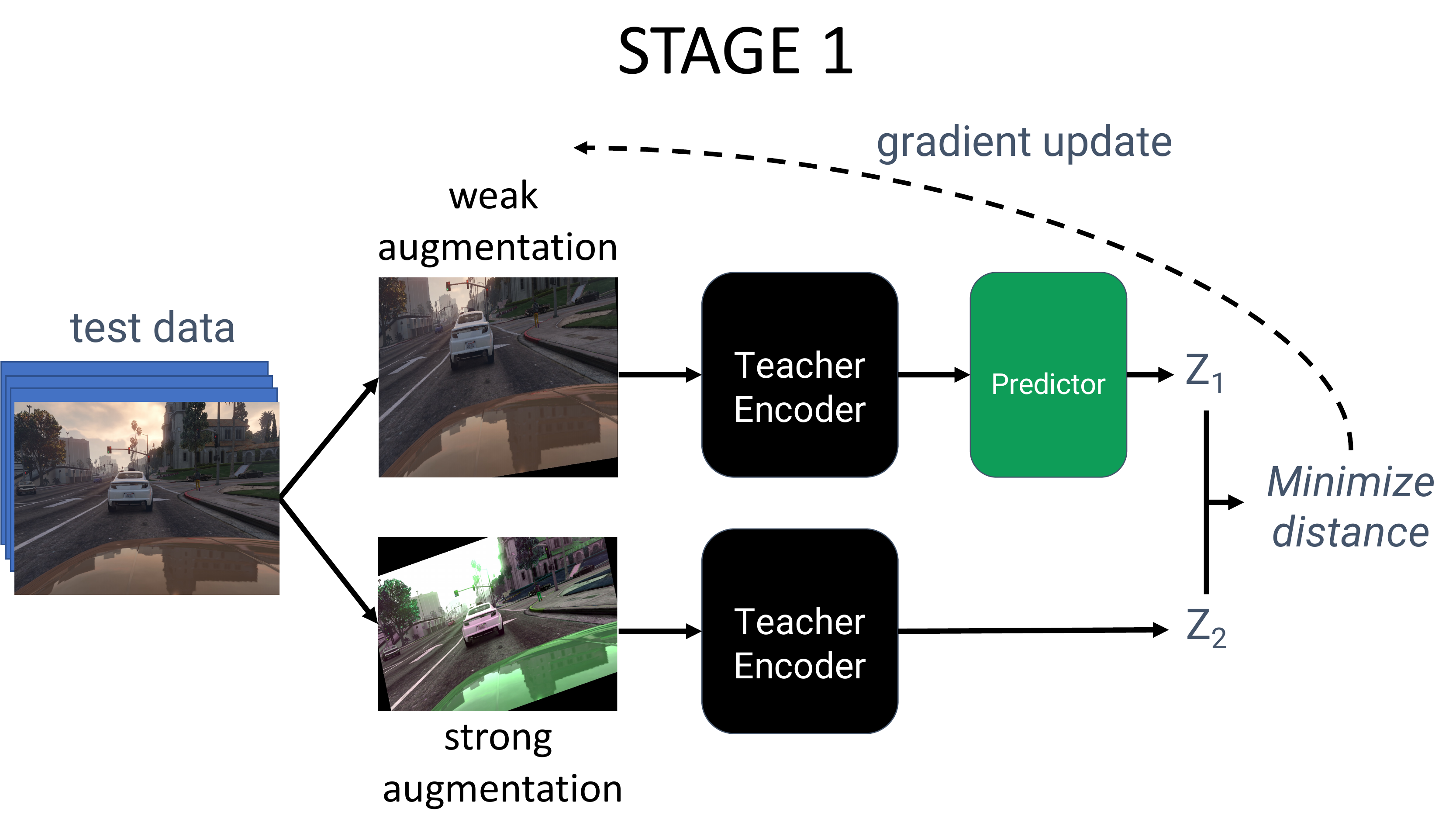}
    \caption{Stage 1: Adapt the \textit{teacher}.}
    \label{subfig:stage1}
    \end{subfigure}
    \begin{subfigure}{0.49\textwidth}
    \includegraphics[width=\textwidth]{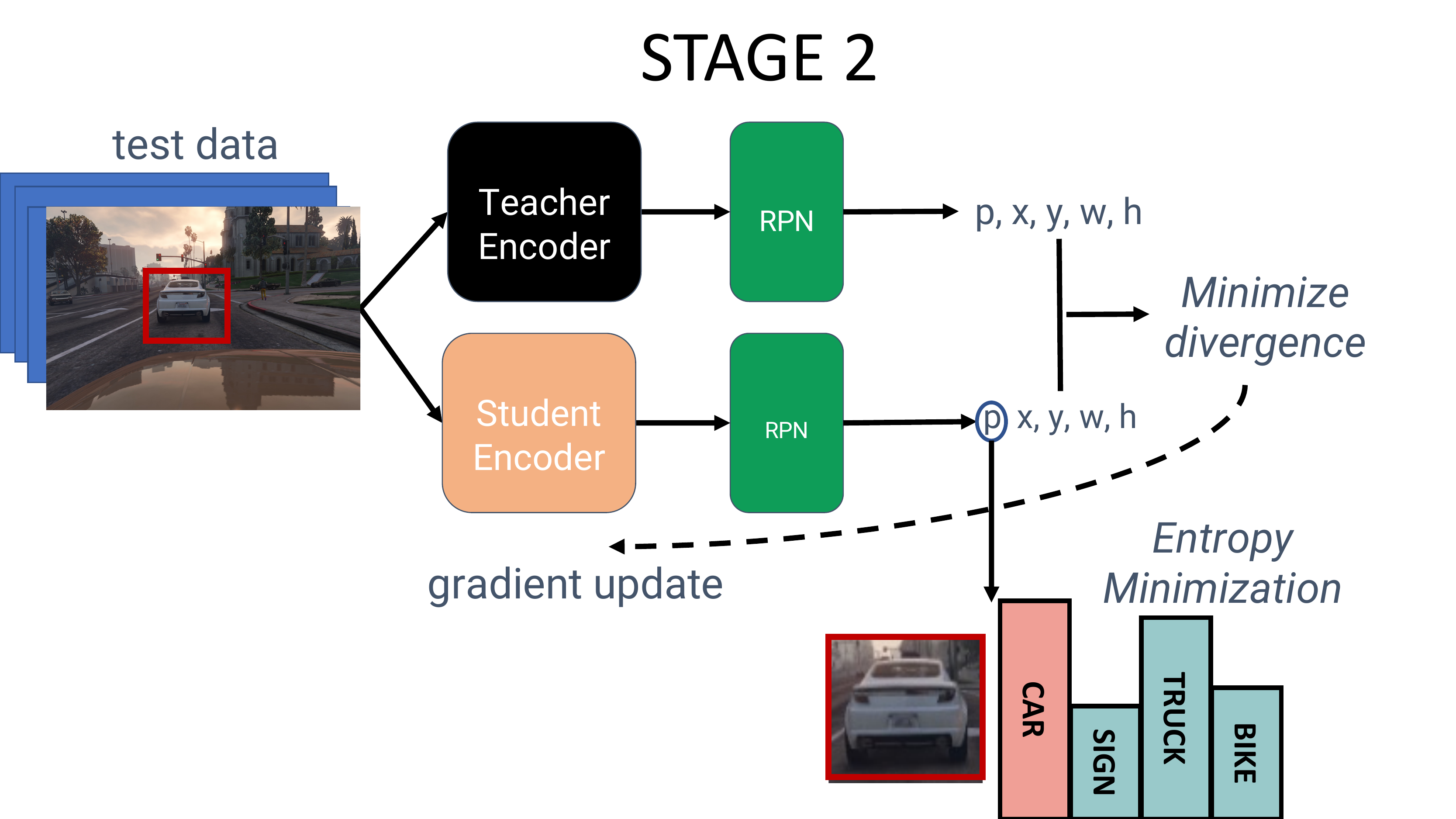}
    \caption{Stage 2: Adapt the \textit{student}.}
    \label{subfig:stage2}
    \end{subfigure}
    \caption{Overview of the Test-time Self-Training architecture (TeST). The model consists of a two-stage
    training process. We first (a) we train a \textit{teacher} network using consistency regularization to learn 
    invariant representations over the novel data distribution. Then (b) we use the trained teacher to distill
    the predictions onto the student).
    }
    \label{fig:teaser}
\end{figure*}
\looseness=-1TeST is a two stage procedure to perform 
test-time adaptation to account for distribution shift.
We propose to finetune a trained \textit{source} model using a two-stage self-training procedure
which first trains the \textit{teacher} using consistency regularization on the novel data distribution 
to learn invariant representations.
Then we leverage knowledge distillation using the teachers learned representations onto the student;
the final student is then evaluated on its ability to generalize to the novel test data.
An overview of the method is presented in Figure \ref{fig:teaser}.
We thoroughly evaluate our method and all baselines on large-scale self-driving object detection and 
image segmentation domain adaptation benchmarks, where we are able to significantly outperform
all baselines.
We observe that TeST consistently outperforms all other test-time adaptation methods
on small and large \textit{budgets} of test data available for finetuning.
We perform our analysis using 6 challenging domain adaptation tasks for object detection,
and 2 challenging tasks for image segmentation.
In our results we find that, TeST significantly outperforms other test-time adaptation algorithms, 
and performs on par with domain adaptation algorithms \cite{chen2020harmonizing,saito2019strong}, 
while requiring 5-10 times fewer target images.

\looseness=-1Our main contributions can be summarized as:
%
(1) We propose a simple test-time adaptation system which utilizes self-distillation to learn robust and 
    invariant representations such that we can effectively adapt to a novel data distributions at test-time.
(2) We exhaustively evaluate the algorithm over 8 challenging detection and segmentation 
    benchmarks and evaluate its effectiveness compared to previous domain adaptation algorithms
    and other test-time training methods, and show that the method is 5-10x more data efficient.

\section{Related Work}
\label{sec:related_work}

\paragraph{Domain adaptation.}
\looseness=-1In domain adaptation a model is given labeled data from a source data distribution and unlabeled 
(or few-labeled) data 
from a target data distribution, and the goal is to learn features such that the model is able to adapt
to the novel data distribution.
Such algorithms rely on learning invariant representation that learn the structure of the task such that
as long as the semantics of the data remains unchanged, the model should be able to generalize to
shifts in distribution. 
Classical domain adaptation algorithms work by kernel mean matching \cite{gretton2009covariate},
by matching metric spaces for SVMs \cite{saenko2010adapting}, or minimizing the moments of distribution
of a learned representation space between the source and target data \cite{tzeng2014deep}.
More recent algorithms perform similar minimization techniques using adversarial optimization \cite{adda},
cyclic losses \cite{hoffman2017cycada}, using coupled generative models \cite{cogan}, or by learning 
joint representation spaces \cite{long2017deep}.
The main challenge with such methods is to learn representations that learn the semantics of the image
which being invariant to exogenous variables, such as weather conditions or noise \cite{zhao2019learning}.

To improve the representation learning over the source and target domains, domain adaptation methods broadly
fall into three categories: methods that try to generate
more data by training class-conditioned generative models \cite{sankaranarayanan2018generate,zhou2020learning},
methods that minimize the distance in the representation space by proposing novel domain
adaptation objectives \cite{shen2018wasserstein,lee2019sliced,adda},
or by adding a regularization to the learned space \cite{uniform_priors,cicek2019unsupervised,mao2019virtual}.
The current state-of-the art algorithms that perform domain adaptation for object detection utilize
decoupling local and global feature alignment \cite{saito2019strong}, maximize discriminability of the 
learned representations along with the transferability \cite{chen2020harmonizing}, use a combination 
of adversarial and reconstruction based approaches \cite{kim2019diversify}, among others 
\cite{xu2020exploring,xu2020cross,zhang2019synthetic,shen2019scl}.
To be able to learn such transferable features, classical domain adaptation algorithms require
access to the data from the target distribution at training time.
This limitation significantly hampers the practical use of such models, since the models can 
only generalize to target distributions that they have encountered during training. 

\paragraph{Test-time adaptation.}
Test-time Training (TTT) was a paradigm that was recently proposed by \cite{ttt}, which trained a model jointly 
on the task loss (image classification) and an image-rotation prediction task during training \cite{rotnet}. 
At test-time, the model was adapted to solve the image rotation task of the test images from a different
distribution. 
The paper argues that during training, solving the image rotation task couples the gradients of the rotation 
prediction task and the task itself, such that performing gradient steps on the novel data over
image rotation prediction objective can help with generalization.
Test-time entropy minimization (Tent) built upon TTT, by proposing to minimize the entropy of the predictions
over the test-data. 
Tent performed gradient steps to minimize the entropy of the models predictions.
Similarly, \cite{mummadi2021test} proposes to maximize the confidence instead of minimizing the entropy of the
predictions, using a confidence heuristic.
More recently, Tailoring \cite{tailoring} was proposed to directly encode inductive biases using a meta-learning
scheme at test time.

Lastly, we note that there was recently a concurrent work that also proposes to perform
test-time adaptation using a student-teacher framework \cite{wang2021target}.
The main differences between TeST and \cite{wang2021target} are that (1) they propose to use
an InfoMax loss \cite{gomes2010discriminative} to train the teacher as opposed to consistency regularization and FixMatch~\cite{fixmatch}; (2) they propose a contrastive learning loss to initialize the student that is re-trained from scratch on the target data as opposed to directly updating the available pre-trained student network. As a consequence, (3) they require significantly more examples despite the simpler problem setting of object recognition. Instead, we tackle challenging computer vision tasks such as image segmentation and object detection, using only a small batch of examples (from 64 to 512).

\paragraph{Self-Training and Student-Teacher framework.}
Student-teacher and self-training algorithms have recently gained popularity due to their ability to 
learn robust representations by transferring information from a \textit{teacher} to a student 
\textit{student} using knowledge distillation \cite{kd}.
Such algorithms have been proposed for semi-supervised learning \cite{tarvainen2017mean}, 
object detection \cite{tang2021humble,xu2021end}, 
image classification \cite{beyer2021knowledge}, few shot learning \cite{ye2021few},
deep metric learning \cite{roth2021simultaneous}, among others.
Recent study in \cite{xie2020self} shows how self-distillation with appropriate data augmentation techniques
can result in state-of-the-art results on the ImageNet classification benchmark.

\section{Preliminaries}
\label{sec:method}





\paragraph{Test-time Adaptation.}
In this paper we consider the test-time adaptation setting, where a model is trained on
the source data $x_\mathcal{S} \sim \mathcal{P}_\mathcal{S}$ but then applied to 
test data $x_\mathcal{T} \sim\mathcal{P}_\mathcal{T}$ from a potentially different distribution, that is
 $\mathcal{P}_\mathcal{S}\neq\mathcal{P}_\mathcal{T}$. The label space is assumed to be
 the same. During source model training we only have access to pairs from the source domain $(x, y) \sim X_{\mathcal{S}}$ and a 
 model is trained using a cross-entropy loss for semantic segmentation, or a sum of cross-entropy and 
regression loss for object detection.
We denote a model trained only on the source data as $\theta_\mathcal{S}$.

\paragraph{Self-Training.}
In a typical self-training framework, a teacher is trained on the labels from the dataset, 
while the student is trained on the pseudo-labels from the teacher's predictions on the data 
(the student has no access to the real labels).
The student is typically trained using a knowledge distillation loss from the teacher's 
predictions \cite{kd}; this setup has shown to result in robust predictions for the student.
After training, the student is then used on the test-set to perform inference.

\section{Test-time Self-Training}
\label{sec:method_name}

\looseness=-1We assume that the input to TeST is a model that was already trained on the source data.
Unlike \cite{ttt}, the model does not need an auxiliary loss during training on the source data and can be trained in any way. To solve the problem of test-time adaptation, we adopt a student-teacher framework in a two-stage setup. 
As we do not assume having access to multiple models trained on the source data distribution $P_\mathcal{S}$ (e.g. bootstrapping),
we copy the source model to create a student $f_\mathcal{S}$ and a teacher model $f_\mathcal{T}$, 
thus initially sharing parameters. 

\looseness=-1In the \textbf{first stage} of adaptation, we only train one of the copied models (the teacher)
using consistency regularization to learn invariant representations for the test data using an additional randomly initialized two layer predictor network as in Figure~\ref{subfig:stage1}. 
The teacher network adapts to the new domain to produce useful pseudo-labels to 
the student in the \textbf{second stage}.
After the teacher is trained on the test-distribution, we utilize self-training to keep the teacher fix and 
train the student. The training objective is supervised and uses the test-data and the pseudo-labels from the teacher. This is depicted in Figure~\ref{subfig:stage2}. 
This self-training step transfers the learned invariant representations
from the teacher to the predictions of the student. 
Finally, similar to \cite{tent}, we add a regularizer to minimize the entropy of the students predictions on the testing
distribution which we empirically found to further improve the performance. 
In the following sections, we will describe each component of the proposed method.

\subsection{Stage 1: Training the teacher model}

The teacher will be used to produce pseudo-labels for the student training. At this point we 
have no supervised signal but just samples from the test distribution, so the goal of training is to learn invariant representations which we hope will translate to mearningful pseudo-labels.
We implement this by consistency regularization 
and FixMatch \cite{fixmatch} as shown in Figure~\ref{subfig:stage1}. We take a \textit{weak} and a \textit{strong} augmentation of the same image from 
the target distribution 
to learn representations 
using extreme examples to enforce stronger invariances.
In practice, for the weak augmentations we use simple transformations such as random rotations,
translations and crops.
As strong augmentation we chose RandAugment \cite{randaugment}, which is trained to find the best 
augmentations for the target task.
RandAugment is also a fast alternative to other automatic augmentation variants,
and therefore can be applied directly at test time to generate strong augmentations
for training the teacher.

After obtaining the features for the strong and weakly augmented images, we use a 
\textit{predictor} network 
$p_\phi$, which is a two-layer ReLU-MLP network, randomly
initialized during the beginning of test-time adaptation following prior work on self-supervised contrastive learning~\cite{byol}. 
Using the teacher $f_{\mathcal{T}}$ and the predictor network $p_\phi$, we 
perform consistency regularization in the feature space to minimize the 
distance between the encoded representations as shown in Figure \ref{subfig:stage1}.
Unlike \cite{fixmatch,crvae}, we minimize the $\ell_2$ distance between the 
strong and the weak embedding on few samples from the target distribution
\begin{equation}
\label{eqn:teacher}
   \min_{\substack{\theta_{\mathcal{T}}} }\bigg|\bigg|f_{\mathcal{T}}\big(strong(x)\big) -p_{\phi}\circ f_{\mathcal{T}}\big(weak(x)\big)\bigg|\bigg|_2^2,
\end{equation}

where $strong(\cdot)$ and $weak(\cdot)$ are strong and weak augmentation policies respectively and $\theta_{\mathcal{T}}$ are the teacher parameters.
Learning invariant representations is key to adaptation to the novel test distribution.
As our main focus are object detection and segmentation, we remark that stage 1 only updates the feature encoder of the teacher networks. 

\subsection{Stage 2: Training the student model}

After training the teacher model, we discard the predictor network $p_\phi$ since it is 
not required to obtain the task predictions.
Using the invariant features from the teacher model, we then train the student using 
pseudo-labels from the teacher, implicitly distilling information about the target distribution as shown in Figure~\ref{subfig:stage2}. 
More specifically, given a batch of test images, we use the teacher model to generate pseudo-labels $\tilde{y}$ and use a knowledge distillation objective to fine-tune the student's 
predictions.
Using knowledge distillation \cite{kd} from the teacher to the student results in robust 
representations learned by the student as the representations learned by the teacher are implicitly 
learned by the student. The per-image $x$ (on the target distribution) training objective reads
\begin{equation}
\label{eqn:KD}
   \min_{\theta_{\mathcal{S}}} \mathcal{L}_{KD}(f_{\mathcal{S}}(x), f_{\mathcal{T}}(\tilde{y}|x))
\end{equation}
where $\mathcal{L}_{KD}$ is the knowledge distillation loss, $f_{\mathcal{S}}$ is the student network parameterized
by $\theta_{\mathcal{S}}$. We note that the loss is only minimized over the students parameters $\theta_{\mathcal{S}}$, and the teacher $f_{\mathcal{T}}$ is kept fixed.
%
\begin{algorithm}[t!]\captionsetup{labelfont={sc,bf}, labelsep=newline}
    \textbf{Input:} Teacher steps $M$;  student steps $N$; \\
    \hspace*{0.35in} Source model $f_{source}$; batch of test data $x_{\mathcal{T}}^{m}$ \\  
    \textbf{Output:} Student model parameters $\theta_{\mathcal{S}}$.
    \begin{algorithmic}[1]
        \State $f_{\mathcal{T}} \gets f_{source}$; \; $f_\mathcal{S} \gets f_{source}$; \;$i,j\gets 0$
        
        \label{alg:main}
        \While {$i < M$}
            \State $\mathcal{L}_{\mathcal{T}} \gets \bigg|\bigg|f_{\mathcal{T}}\big(strong(x)\big) - p_{\phi}\circ f_{\mathcal{T}}\big(weak(x)\big)\bigg|\bigg|_2^2$
            \State $\theta_{\mathcal{T}} \gets \theta_{\mathcal{T}} - \alpha \nabla \mathcal{L}_{\mathcal{T}}$ 
            \State $i \gets i + 1$
        \EndWhile
        
        \While{$j < N$}
            \State $\tilde{y} \gets f_{\mathcal{T}}(x_{\mathcal{T}})$
            \State $\mathcal{L}_{\mathcal{S}} \gets \mathcal{L}_{KD}(f_{\mathcal{S}}(x_{\mathcal{T}}), \tilde{y}) + \lambda \mathcal{H}(f_{\mathcal{S}}(x_{\mathcal{T}}))$ 
            \State $\theta_{\mathcal{S}} \gets \theta_{\mathcal{S}} - \alpha \nabla \mathcal{L}_{\mathcal{S}}$ 
            \State $j \gets j + 1$
        \EndWhile
    \end{algorithmic}
\end{algorithm}
The knowledge distillation loss used for a categorical variable (such as bounding box classification) is the 
KL-Divergence between the two probability distributions, and for a continuous variable (such as bounding
box regression) is the L2 distance between the pseudo-labels and the student labels.

To improve the confidence of the student predictions, we add an entropy minimization term over the probability 
distribution of the class predictions, as confidence has been linked to performance \cite{tent}.
The entropy minimization term is added as a regularization to the student training.
Adding a weighted entropy term to the objective in Equation \ref{eqn:KD}, the full student loss is thus:
\begin{equation}
\label{eqn:full}
    \min_{\theta_{\mathcal{S}}} \mathcal{L}_{KD}(f_{\mathcal{S}}(x), f_{\mathcal{T}}(\tilde{y}|x))+\lambda \mathcal{H}(f_{\mathcal{S}}(x)),    
\end{equation}

where $\lambda$ is a hyperparameter that weights the entropy term $\mathcal{H}$ sharpening the confidence of the class prediction.

\subsection{Concluding remarks}

To summarize, we adapt a pre-trained source model $f_{\mathcal{S}}$ in two stages to
the novel test distribution $\mathcal{P}_{\mathcal{T}}$: first we train the teacher  
by minimizing the consistency regularization loss using strong and weak augmentations of the target images, 
and secondly we use the trained teacher to provide pseudo-labels for the student to train on. 
The student is then optimized using gradient descent over the weighted knowledge distillation and entropy 
minimization loss, as in Eqn. \ref{eqn:full}, see Section \ref{sec:exp-setup} for experimental details.
The joint training algorithm can be found in Algorithm \ref{alg:main}.
This procedure is generic and can be used for many computer vision domain adaptation task. In the following we 
test the performance of TeST on object detection and image segmentation tasks.


\section{Experiments}
\label{sec:exps}

In this section, we seek to answer the following questions:
\begin{itemize}
    \item Can TeST yield better domain adaptation performance on large-scale object detection tasks than baseline test-time adaptation methods? 
    How does it compare against domain adaptation methods where test data is available during training time?
    \item Can TeST yield better domain adaptation performance on further computer vision tasks, such as image segmentation?
    \item Is TeST robust to our assumption of distribution shift and performs better representation learning even if the testing distribution is the same as the training
    distribution? 
\end{itemize}

\subsection{Experimental setup}
\label{sec:exp-setup}

\paragraph{Test data budget:} In test-time adaptation, the model is able to adapt to a certain \textit{budget} $n$ of images available during test-time 
to adapt to the novel data distribution.
To tho\-rough\-ly examine our proposed solution, we test over a range of values for $n\in\{64,128,256,512\}$.

\paragraph{Baselines:} We consider a range of baselines. \textbf{Source Only}: refers to training on source data
$X_\mathcal{S}$ only and no further adaptation.
In addition, we use two recent test-time adaptation models as baselines for all object detection experiments: \textbf{Test-Time Training} (TTT) 
which proposes to solve an image rotation prediction task at training time and test time to adapt to novel distributions 
\cite{ttt}. 
\textbf{Tent}: which proposes to minimize the entropy of the test predictions as an unsupervised task at test-time \cite{tent}.
Furthermore, we also show results with the model being trained with access to the test-distribution without labels
during training time as considered in unsupervised domain adaptation setting.
\textbf{SHOT}: which proposes to use a self-supervised objective to maximize the alignment of the features between the original source data and the novel target distribution \cite{shotpp}.
Namely, we use two recently proposed unsupervised domain adaptation algorithms \textbf{Saito et al.} \cite{saito2019strong} and \textbf{Chen et al.} \cite{chen2020harmonizing}. 
Finally, we also compare our results with a \textbf{Finetuning} oracle, where we consider the performance of the trained model on the test-set if 
it had access to $n$ labeled samples. Although it is possible to improve the \textit{fine-tune} results using more recent transfer learning algorithms \cite{spottune,stochnorm,delta}, we update all parameters of the network trained on the source domain using the ground-truth label information as opposed to the pseudo-labels in the knowledge distillation loss of Equation~\ref{eqn:full} (without the entropy regularization).
\begin{figure*}[t!]
    \centering
    \begin{subfigure}{0.24\textwidth}
    \includegraphics[width=\textwidth]{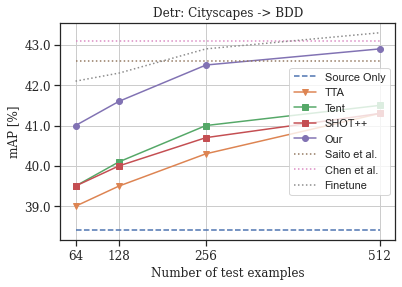}
    \caption{Cityscapes $\rightarrow$ BDD100k}
    \end{subfigure}
    \begin{subfigure}{0.24\textwidth}
    \includegraphics[width=\textwidth]{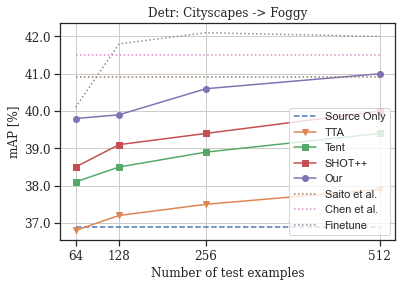}
    \caption{Cityscapes $\rightarrow$ Foggy}
    \end{subfigure}
    \begin{subfigure}{0.24\textwidth}
    \includegraphics[width=\textwidth]{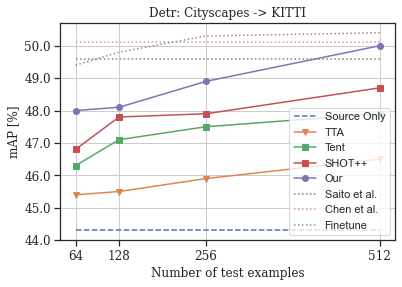}
    \caption{Cityscapes $\rightarrow$ KITTI}
    \end{subfigure}
    \begin{subfigure}{0.24\textwidth}
    \includegraphics[width=\textwidth]{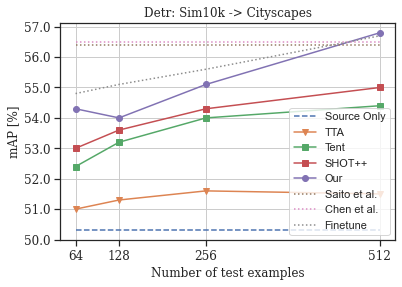}
    \caption{Sim10k $\rightarrow$ Cityscapes}
    \end{subfigure}
    \caption{Results on object detection with self-driving datasets where the base detector is a Deformable Detection Transformer is used \cite{deformable_detr}.
    We see that we are able to significantly outperform other test-time adaptation models (solid lines), while being comparable 
    to domain adaptation models \cite{chen2020harmonizing} (brown dotted line) and  \cite{saito2019strong} (violet dotted line)
    which require 5-10x more data from the target data distribution.
    TeST is also comparable to the \textbf{Finetuning} baseline, which directly finetunes on the budget of ground-truth
    labels on the test distribution, instead of using an unsupervised objective.
    }
    \label{fig:detr-driving-detection-results}
\end{figure*}

\begin{figure*}[t!]
    \centering
    \begin{subfigure}{0.24\textwidth}
    \includegraphics[width=\textwidth]{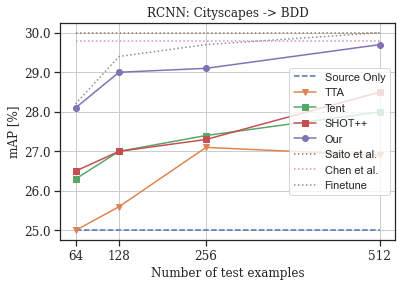}
    \caption{Cityscapes $\rightarrow$ BDD100k}
    \end{subfigure}
    \begin{subfigure}{0.24\textwidth}
    \includegraphics[width=\textwidth]{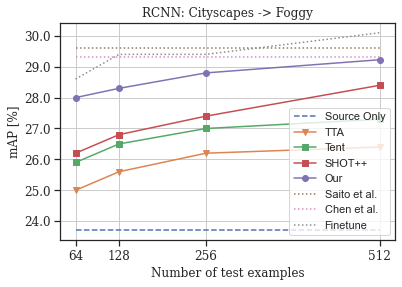}
    \caption{Cityscapes $\rightarrow$ Foggy}
    \end{subfigure}
    \begin{subfigure}{0.24\textwidth}
    \includegraphics[width=\textwidth]{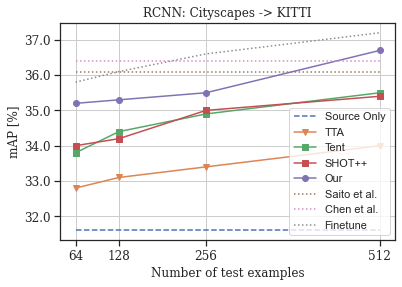}
    \caption{Cityscapes $\rightarrow$ KITTI}
    \end{subfigure}
    \begin{subfigure}{0.24\textwidth}
    \includegraphics[width=\textwidth]{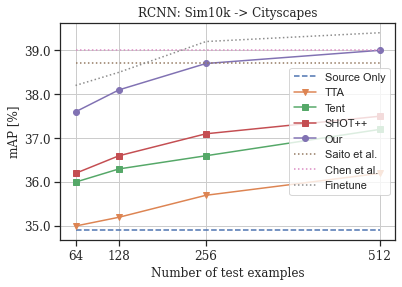}
    \caption{Sim10k $\rightarrow$ Cityscapes}
    \end{subfigure}
    \caption{Results on object detection with self-driving datasets with a Faster RCNN base object detector \cite{ren2015faster}.
    TeST significantly outperforms the each of the test-time adaptation baselines (solid lines), and is comparable
    and often better than the domain adaptation baselines 
    \cite{chen2020harmonizing} (brown dotted line) and  \cite{saito2019strong} (violet dotted line).
    TeST is also able to obtain similar performance to the \textit{oracle} baseline, which finetunes on the labeled
    data points, instead of using an unsupervised objective.}
    \label{fig:rcnn-driving}
\end{figure*}

\paragraph{Architectures:} For the object detection experiments, we use two  architectures: Faster RCNN \cite{ren2015faster} and Deformable Detection
Transformer \cite{deformable_detr} (DeTR). We train them on the source datasets using the default hyperparameters mentioned in the 
original papers.
For image segmentation, we train a Dilated Residual Network \cite{drn} on the source data again using default hyperparameters.

\paragraph{Datasets:} We consider a variety of self-driving object detection domain adaptation challenges, namely:
Cityscapes \cite{cityscapes} $\rightarrow$ BDD \cite{bdd}, Cityscapes $\rightarrow$ Foggy Cityscapes \cite{foggy},
Cityscapes $\rightarrow$ KITTI \cite{kitti} and Sim10k \cite{sim10k} $\rightarrow$ Cityscapes.
This set of experiments measure a variety of distribution shifts, such as weather, location and sim-2-real.
Additionally, we also generate OOD splits from MS-COCO \cite{coco} 
by extracting the features using a trained ResNet-101 model \cite{resnet}, and then 
 generating $k$-clusters of the embedding to separate the data along some unknown visual or semantic 
 boundaries.
The motivation is twofold. First, we want to test the adaptation algorithms on \textit{common objects} as in the ones in MS-COCO. Second, generating $k$-clusters and varying $k$ we can control the severity of the distribution shift between training and test time and evaluate how this affects performance of the tested approaches.
We hold out one cluster and use it as a \textit{OOD} target distribution, training the model on the $k-1$ clusters. 
Smaller $k$ corresponds to more severe shift as some object classes may even be missing from the source training distribution. 
However, manually checking the label distribution of the clusters, confirmed 
that at least a few examples of each of the classes are present in each cluster.
More details regarding all test-settings and datasets can be found in Appendix \ref{app:tasks}.

\paragraph{Hyperparameters:} To train TeST, we use an Adam optimizer \cite{Adam} with a learning rate of $3 \times 10^{-4}$ for
both the student and teacher, and perform 10 epochs over the $n$-budget for each of the stages.
As mentioned previously, to generate \textit{strong} augmentations, we use a RandAugment policy \cite{randaugment};
to generate \textit{weak} augmentations, we perform rotations from $[-10^{\circ}, 10^{\circ}]$, and random crops
of the original image.
For all experiments, we use an entropy weight $\lambda=0.25$.

\begin{figure*}[t!]
    \centering
    \begin{subfigure}{0.24\textwidth}
    \includegraphics[width=\textwidth]{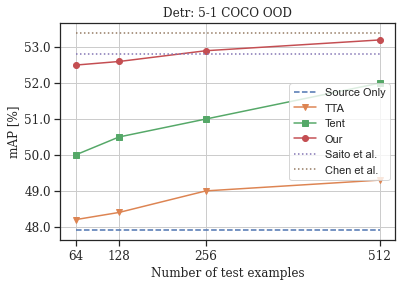}
    \caption{DeTR: 4 train - 1 test}
    \end{subfigure}
    \begin{subfigure}{0.24\textwidth}
    \includegraphics[width=\textwidth]{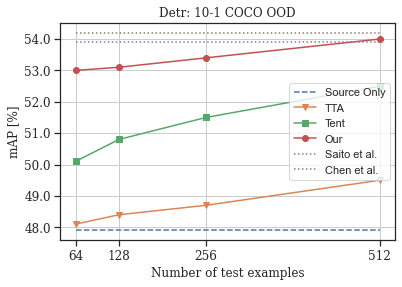}
    \caption{DeTR: 9 train - 1 test}
    \end{subfigure}
    \begin{subfigure}{0.24\textwidth}
    \includegraphics[width=\textwidth]{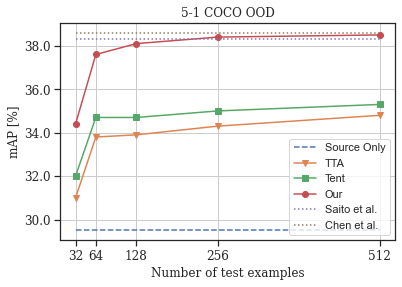}
    \caption{RCNN: 4 train - 1 test}
    \end{subfigure}
    \begin{subfigure}{0.24\textwidth}
    \includegraphics[width=\textwidth]{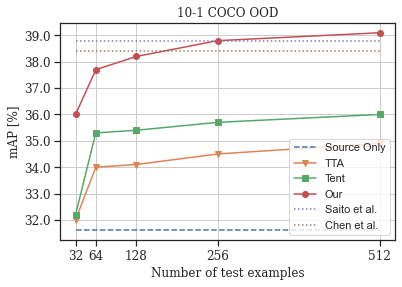}
    \caption{RCNN: 9 train - 1 test}
    \end{subfigure}
    \caption{Results on object detection using the MS-COCO dataset with Deformable DeTR and Faster RCNN.
    We cluster the MS-COCO dataset using the embeddings of a pretrained ResNet feature extractor into $k$-clusters.
    We then randomly select one cluster to be the out-of-distribution test set, and use 
    the remaining $k-1$ clusters for training. 
    This way we are able to evaluate domain adaptation algorithms on \textit{unknown} distribution shifts,
    as the clusters are separated over an unknown semantic boundary.
    We continue to see that TeST is able to outperform the test-time adaptation baselines (solid lines)
    and is competitive with unsupervised domain adaptation models (dotted lines) which are trained with 
    data from all $k$-clusters. 
    }
    \label{fig:coco-detection-results}
\end{figure*}

\begin{figure*}[t!]
    \centering
    \begin{subfigure}{0.34\textwidth}
    \includegraphics[width=\textwidth]{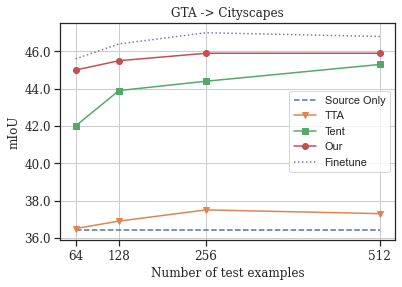}
    \caption{GTA $\rightarrow$ Cityscapes}
    \end{subfigure}
    \begin{subfigure}{0.34\textwidth}
    \includegraphics[width=\textwidth]{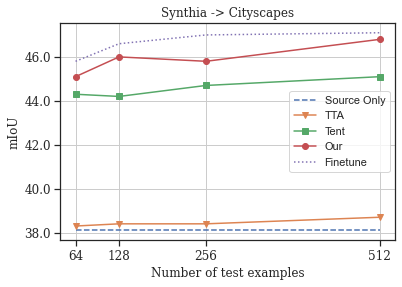}
    \caption{Synthia $\rightarrow$ Cityscapes}
    \end{subfigure}
    \caption{Results on image segmentation with DRN over two challenging Sim2Real driving tasks.
    We compare TeST to 2 test-time adaptation algorithms, and directly finetuning, and see that
    TeST is comparable to finetuning on the labels directly.}
    \label{fig:drn-segmentation-results}
\end{figure*}

\subsection{Results}

\paragraph{Self-Driving Object Detection Benchmarks} 
We test over a variety of benchmark self-driving object detection datasets and their domain adaptation variants which include change in weather (Cityscapes
$\rightarrow$ Foggy Cityscapes), change in location (Cityscapes $\rightarrow$ KITTI and Cityscapes $\rightarrow$ BDD) and from simulated to real data (Sim10k $\rightarrow$ Cityscapes).
The results for Deformable DeTR are shown in Figure \ref{fig:detr-driving-detection-results}, and the results for
FasterRCNN can be found in Figure \ref{fig:rcnn-driving}.
In both cases, we see that TeST is able to significantly outperform both the test-time adaptation baselines and
the model that was only trained on source data (\textit{Source Only} in the plots).
Interestingly, we note that TeST is able to achieve performance comparable to unsupervised domain adaptation
algorithms from Chen et al.\cite{chen2020harmonizing} and Saito et al. \cite{saito2019strong}
 despite only using between 256 and 512 test-images. Both of the domain adaptation algorithms
were trained with the whole training dataset of the target tasks (see Appendix \ref{app:tasks} for further details).
This results in TeST performing comparably to such algorithms, while being 5-10x more data efficient, as TeST only
requires 256-512 unlabeled images to obtain similar performance.
This suggests that instead of assuming that all source images are available at training time, like in 
\cite{saito2019strong,chen2020harmonizing},
test-time adaptation provides a data-efficient alternative.
We also note that TeST performs comparably at times to the oracle \textit{finetuning} baseline where the model is finetuned on $n$ labeled examples, as opposed to having to perform unsupervised adaptation at test-time.

\paragraph{Object Detection on MS-COCO}
MS-COCO is a versatile benchmark dataset for several computer vision tasks including object detection.
To test domain adaptation on MS-COCO we split the data in $k$ clusters based on the trained embeddings of a 
ResNet-101 model. Training is then done on $k-1$ clusters, the held-out cluster is used for testing.
This clustering technique separates creates an OOD test-set with an unknown domain shift.
Improving the performances on unknown domain shifts is important to ensure that the models did not 
overfit to commonly tested distribution shifts, such as weather, location, an sim-2-real.
We detail the dataset splitting further in Appendix \ref{app:tasks}.
In practice, we vary the total number of clusters, $k\in\{5, 10\}$, which means 4 or 9 clusters
are available for training, respectively, while 1 is held out for testing.
The fewer clusters, the larger the expected distribution shift between the training
and test distributions.
As before, we train a Faster-RCNN object detector \cite{ren2015faster} and a Deformable 
DeTR \cite{deformable_detr} on the training clusters, and then evaluate the chosen
test-time adaptation baselines, and TeST on the testing cluster.

Full results on MS-COCO are present in Figure \ref{fig:coco-detection-results}.
Similar as before, we see considerable improvements compared to the test-time baselines, while again achieving
comparable performance to the domain adaptation algorithms despite using significantly less data. 
As expected, we observe that the performance is overall slightly worse for more severe shifts (fewer clusters). However, this has no effects on the conclusions we can draw, confirming the trends we observed on the other datasets.
This further suggests that TeST is able to get empirical gains across different domains
(self-driving datasets and common objects) across two popular object detectors and severity of the shift.

\paragraph{Semantic Segmentation for Self-Driving}
To test the generality of TeST, it is important for the algorithm to be agnostic
to the type of task considered. 
Towards this, we further perform experiments with semantic segmentation on self-driving datasets.
We consider the same test-time setting as before, and use the popular Sim2Real benchmarks of
GTA $\rightarrow$ Cityscapes and Synthia $\rightarrow$ Cityscapes \cite{synthia}, where the model is trained
on simulation data, and tested on the natural-image Cityscapes dataset \cite{cityscapes}.
The model is trained using a Dilated Residual Network \cite{drn} using the default hyperparameters as
suggested in the original paper.
\begin{figure*}[t!]
    \centering
    \begin{subfigure}{0.34\textwidth}
    \includegraphics[width=\textwidth]{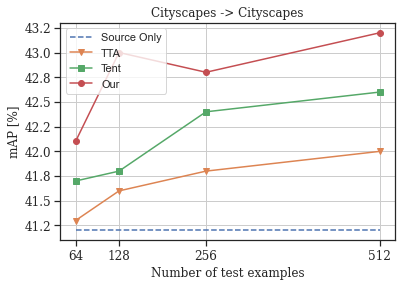}
    \caption{Faster-RCNN}
    \end{subfigure}
    \begin{subfigure}{0.34\textwidth}
    \includegraphics[width=\textwidth]{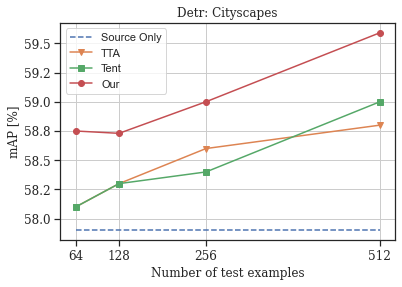}
    \caption{Deformable DeTR}
    \end{subfigure}
    \caption{Results on object detection for Cityscapes (no OOD) using Faster RCNN (left) and Deformable DeTR (right).
    The results suggest that even when there is no distribution shift between the training and the testing distribution,
    TeST is able to outperform other test-time adaptation baselines.
    This further shows that TeST is robust to any or no distribution shift.}
    \label{fig:cityscapes-no-ood-results}
\end{figure*}

\begin{figure}
    \centering
    \includegraphics[width=0.34\textwidth]{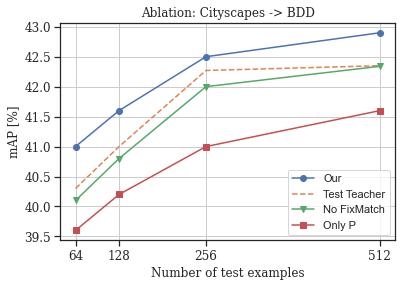}
    \caption{Ablation study over different possible variants of TeST on the benchmark Cityscapes $\rightarrow$ BDD domain adaptation challenge.}
    \label{fig:ablation}
\end{figure}
The results are shown in Figure \ref{fig:drn-segmentation-results}.
Again, we see that TeST is able to outperform the baseline algorithms by a considerable margin,
which suggests that the proposed method is indeed task-agnostic and applies beyond detection only..
Interestingly, we see that Tent \cite{tent} significantly outperforms the Test-Time Training baseline (TTT)
which also shows the importance of entropy minimization at test-time, which is also a component of TeST.

\looseness=-1\paragraph{Investigating No Distribution Shift} TeST assumes that the target distribution is different than the source distribution. This raises the natural question of what happens whenever there is actually \textbf{no} distribution
shift between the training and the testing distribution. Our goal is to understand whether TeST is robust to violation of this assumption as in practical scenarios it may be difficult to know for certain that the distribution has changed. 
Prior work found that self-distillation improves performance on the training dataset~\cite{furlanello2018born}. While our approach was designed to cope with distribution shifts, we would still hope to see gains even if the distribution did not change.
For this, we use the original test set of the source dataset itself.
Similar to before, we use the Cityscapes dataset \cite{cityscapes} for training, and evaluate on the held-out
validation set. 

The results for a Faster-RCNN and Deformable DeTR detectors are in Figure \ref{fig:cityscapes-no-ood-results} and continue to show the broad applicability of TeST: even if the distribution did not change, we are able to significantly improve performance.


\looseness=-1\paragraph{Ablation Study}
Finally, we present different variations of TeST and report their performance on
the Cityscapes $\rightarrow$ BDD object detection benchmark using a Deformable
DeTR in Figure \ref{fig:ablation}.
We observe the largest performance decrease if we only perform knowledge distillation on the probability output $p$ of the bounding
box i.e. no bounding box regression (denoted as ``Only p'' in Figure~\ref{fig:ablation}).
Similarly, we can see the benefit of learning invariant representations in the teacher network
using FixMatch and the consistency regularization as opposed to simply doing self-distillation (No FixMatch).
Finally, we also note that it is possible to use the Teacher directly (``Test Teacher'' in 
Figure \ref{fig:ablation}), however, using knowledge distillation clearly helps to learn a robust student. Overall, each component of the method yields a positive contribution to the final performance. 
This further suggests the TeST builds upon several components that work well together, to
perform better representation learning through self-training.

\section{Conclusion}
In this paper, we propose TeST: a test-time adaptation technique that uses
self-training in a student-teacher framework to overcome distribution shifts 
at test-time.
The key ingredients are (1) learning invariant and robust representations of the test distribution, 
and (2) distilling the predictions to the student model.
TeST consists of two-stage, a teacher training through consistency regularization followed by 
knowledge distillation and
entropy minimization to train a student model. 
Overall, TeST significantly outperforms test-time adaption baselines, and is comparable to unsupervised domain adaptation techniques that require 5-10x more data from the target distribution during training.

{\small
\bibliographystyle{ieee_fullname}
\bibliography{egbib}

\begin{thebibliography}{10}\itemsep=-1pt

\bibitem{tailoring}
Ferran Alet, Maria Bauza, Kenji Kawaguchi, Nurullah~Giray Kuru, Tomas
  Lozano-Perez, and Leslie~Pack Kaelbling.
\newblock Tailoring: encoding inductive biases by optimizing unsupervised
  objectives at prediction time.
\newblock {\em arXiv preprint arXiv:2009.10623}, 2020.

\bibitem{sup_da_3}
Shuang Ao, Xiang Li, and Charles Ling.
\newblock Fast generalized distillation for semi-supervised domain adaptation.
\newblock In {\em Proceedings of the AAAI Conference on Artificial
  Intelligence}, 2017.

\bibitem{beyer2021knowledge}
Lucas Beyer, Xiaohua Zhai, Am{\'e}lie Royer, Larisa Markeeva, Rohan Anil, and
  Alexander Kolesnikov.
\newblock Knowledge distillation: A good teacher is patient and consistent.
\newblock {\em arXiv preprint arXiv:2106.05237}, 2021.

\bibitem{chen2020harmonizing}
Chaoqi Chen, Zebiao Zheng, Xinghao Ding, Yue Huang, and Qi Dou.
\newblock Harmonizing transferability and discriminability for adapting object
  detectors.
\newblock In {\em Proceedings of the IEEE/CVF Conference on Computer Vision and
  Pattern Recognition}, pages 8869--8878, 2020.

\bibitem{cicek2019unsupervised}
Safa Cicek and Stefano Soatto.
\newblock Unsupervised domain adaptation via regularized conditional alignment.
\newblock In {\em Proceedings of the IEEE/CVF International Conference on
  Computer Vision}, pages 1416--1425, 2019.

\bibitem{cityscapes}
Marius Cordts, Mohamed Omran, Sebastian Ramos, Timo Rehfeld, Markus Enzweiler,
  Rodrigo Benenson, Uwe Franke, Stefan Roth, and Bernt Schiele.
\newblock The cityscapes dataset for semantic urban scene understanding.
\newblock In {\em Proceedings of the IEEE conference on computer vision and
  pattern recognition}, pages 3213--3223, 2016.

\bibitem{randaugment}
Ekin~D Cubuk, Barret Zoph, Jonathon Shlens, and Quoc~V Le.
\newblock Randaugment: Practical automated data augmentation with a reduced
  search space.
\newblock In {\em Proceedings of the IEEE/CVF Conference on Computer Vision and
  Pattern Recognition Workshops}, pages 702--703, 2020.

\bibitem{furlanello2018born}
Tommaso Furlanello, Zachary Lipton, Michael Tschannen, Laurent Itti, and Anima
  Anandkumar.
\newblock Born again neural networks.
\newblock In {\em International Conference on Machine Learning}, pages
  1607--1616. PMLR, 2018.

\bibitem{kitti}
Andreas Geiger, Philip Lenz, and Raquel Urtasun.
\newblock Are we ready for autonomous driving? the kitti vision benchmark
  suite.
\newblock In {\em 2012 IEEE conference on computer vision and pattern
  recognition}, pages 3354--3361. IEEE, 2012.

\bibitem{gen_1}
Robert Geirhos, Carlos R~Medina Temme, Jonas Rauber, Heiko~H Sch{\"u}tt,
  Matthias Bethge, and Felix~A Wichmann.
\newblock Generalisation in humans and deep neural networks.
\newblock {\em arXiv preprint arXiv:1808.08750}, 2018.

\bibitem{rotnet}
Spyros Gidaris, Praveer Singh, and Nikos Komodakis.
\newblock Unsupervised representation learning by predicting image rotations.
\newblock {\em arXiv preprint arXiv:1803.07728}, 2018.

\bibitem{fast_rcnn}
Ross Girshick.
\newblock Fast r-cnn.
\newblock In {\em Proceedings of the IEEE international conference on computer
  vision}, pages 1440--1448, 2015.

\bibitem{gomes2010discriminative}
Ryan Gomes, Andreas Krause, and Pietro Perona.
\newblock Discriminative clustering by regularized information maximization.
\newblock In {\em Nueral Information Processing Systems}, 2010.

\bibitem{gretton2009covariate}
Arthur Gretton, Alex Smola, Jiayuan Huang, Marcel Schmittfull, Karsten
  Borgwardt, and Bernhard Sch{\"o}lkopf.
\newblock Covariate shift by kernel mean matching.
\newblock {\em Dataset shift in machine learning}, 3(4):5, 2009.

\bibitem{byol}
Jean-Bastien Grill, Florian Strub, Florent Altch{\'e}, Corentin Tallec,
  Pierre~H Richemond, Elena Buchatskaya, Carl Doersch, Bernardo~Avila Pires,
  Zhaohan~Daniel Guo, Mohammad~Gheshlaghi Azar, et~al.
\newblock Bootstrap your own latent: A new approach to self-supervised
  learning.
\newblock {\em arXiv preprint arXiv:2006.07733}, 2020.

\bibitem{spottune}
Yunhui Guo, Honghui Shi, Abhishek Kumar, Kristen Grauman, Tajana Rosing, and
  Rogerio Feris.
\newblock Spottune: transfer learning through adaptive fine-tuning.
\newblock In {\em Proceedings of the IEEE/CVF Conference on Computer Vision and
  Pattern Recognition}, pages 4805--4814, 2019.

\bibitem{mask_rcnn}
Kaiming He, Georgia Gkioxari, Piotr Doll{\'a}r, and Ross Girshick.
\newblock Mask r-cnn.
\newblock In {\em Proceedings of the IEEE international conference on computer
  vision}, pages 2961--2969, 2017.

\bibitem{resnet}
Kaiming He, Xiangyu Zhang, Shaoqing Ren, and Jian Sun.
\newblock Deep residual learning for image recognition.
\newblock In {\em Proceedings of the IEEE conference on computer vision and
  pattern recognition}, pages 770--778, 2016.

\bibitem{kd}
Geoffrey Hinton, Oriol Vinyals, and Jeff Dean.
\newblock Distilling the knowledge in a neural network.
\newblock {\em arXiv preprint arXiv:1503.02531}, 2015.

\bibitem{hoffman2017cycada}
Judy Hoffman, Eric Tzeng, Taesung Park, Jun-Yan Zhu, Phillip Isola, Kate
  Saenko, Alexei~A Efros, and Trevor Darrell.
\newblock Cycada: Cycle-consistent adversarial domain adaptation.
\newblock {\em arXiv preprint arXiv:1711.03213}, 2017.

\bibitem{sup_da_4}
Pin Jiang, Aming Wu, Yahong Han, Yunfeng Shao, Meiyu Qi, and Bingshuai Li.
\newblock Bidirectional adversarial training for semi-supervised domain
  adaptation.
\newblock In {\em IJCAI}, pages 934--940, 2020.

\bibitem{sim10k}
Matthew Johnson-Roberson, Charles Barto, Rounak Mehta, Sharath~Nittur Sridhar,
  Karl Rosaen, and Ram Vasudevan.
\newblock Driving in the matrix: Can virtual worlds replace human-generated
  annotations for real world tasks?
\newblock {\em arXiv preprint arXiv:1610.01983}, 2016.

\bibitem{kim2019diversify}
Taekyung Kim, Minki Jeong, Seunghyeon Kim, Seokeon Choi, and Changick Kim.
\newblock Diversify and match: A domain adaptive representation learning
  paradigm for object detection.
\newblock In {\em Proceedings of the IEEE/CVF Conference on Computer Vision and
  Pattern Recognition}, pages 12456--12465, 2019.

\bibitem{Adam}
Diederik~P Kingma and Jimmy Ba.
\newblock Adam: A method for stochastic optimization.
\newblock {\em arXiv preprint arXiv:1412.6980}, 2014.

\bibitem{stochnorm}
Zhi Kou, Kaichao You, Mingsheng Long, and Jianmin Wang.
\newblock Stochastic normalization.
\newblock {\em Advances in Neural Information Processing Systems}, 33, 2020.

\bibitem{dece}
Fabian Kuppers, Jan Kronenberger, Amirhossein Shantia, and Anselm Haselhoff.
\newblock Multivariate confidence calibration for object detection.
\newblock In {\em Proceedings of the IEEE/CVF Conference on Computer Vision and
  Pattern Recognition Workshops}, pages 326--327, 2020.

\bibitem{lee2019sliced}
Chen-Yu Lee, Tanmay Batra, Mohammad~Haris Baig, and Daniel Ulbricht.
\newblock Sliced wasserstein discrepancy for unsupervised domain adaptation.
\newblock In {\em Proceedings of the IEEE/CVF Conference on Computer Vision and
  Pattern Recognition}, pages 10285--10295, 2019.

\bibitem{delta}
Xingjian Li, Haoyi Xiong, Hanchao Wang, Yuxuan Rao, Liping Liu, Zeyu Chen, and
  Jun Huan.
\newblock Delta: Deep learning transfer using feature map with attention for
  convolutional networks.
\newblock {\em arXiv preprint arXiv:1901.09229}, 2019.

\bibitem{shotpp}
Jian Liang, Dapeng Hu, Yunbo Wang, Ran He, and Jiashi Feng.
\newblock Source data-absent unsupervised domain adaptation through hypothesis
  transfer and labeling transfer.
\newblock {\em IEEE Transactions on Pattern Analysis and Machine Intelligence},
  2021.

\bibitem{fpn}
Tsung-Yi Lin, Piotr Doll{\'a}r, Ross Girshick, Kaiming He, Bharath Hariharan,
  and Serge Belongie.
\newblock Feature pyramid networks for object detection.
\newblock In {\em Proceedings of the IEEE conference on computer vision and
  pattern recognition}, pages 2117--2125, 2017.

\bibitem{coco}
Tsung-Yi Lin, Michael Maire, Serge Belongie, James Hays, Pietro Perona, Deva
  Ramanan, Piotr Doll{\'a}r, and C~Lawrence Zitnick.
\newblock Microsoft coco: Common objects in context.
\newblock In {\em European conference on computer vision}, pages 740--755.
  Springer, 2014.

\bibitem{cogan}
Ming-Yu Liu and Oncel Tuzel.
\newblock Coupled generative adversarial networks.
\newblock {\em Advances in neural information processing systems}, 29:469--477,
  2016.

\bibitem{fcn}
Jonathan Long, Evan Shelhamer, and Trevor Darrell.
\newblock Fully convolutional networks for semantic segmentation.
\newblock In {\em Proceedings of the IEEE conference on computer vision and
  pattern recognition}, pages 3431--3440, 2015.

\bibitem{long2017deep}
Mingsheng Long, Han Zhu, Jianmin Wang, and Michael~I Jordan.
\newblock Deep transfer learning with joint adaptation networks.
\newblock In {\em International conference on machine learning}, pages
  2208--2217. PMLR, 2017.

\bibitem{mao2019virtual}
Xudong Mao, Yun Ma, Zhenguo Yang, Yangbin Chen, and Qing Li.
\newblock Virtual mixup training for unsupervised domain adaptation.
\newblock {\em arXiv preprint arXiv:1905.04215}, 2019.

\bibitem{gen_2}
Timo Milbich, Karsten Roth, Samarth Sinha, Ludwig Schmidt, Marzyeh Ghassemi,
  and Bj{\"o}rn Ommer.
\newblock Characterizing generalization under out-of-distribution shifts in
  deep metric learning.
\newblock {\em arXiv preprint arXiv:2107.09562}, 2021.

\bibitem{mummadi2021test}
Chaithanya~Kumar Mummadi, Robin Hutmacher, Kilian Rambach, Evgeny Levinkov,
  Thomas Brox, and Jan~Hendrik Metzen.
\newblock Test-time adaptation to distribution shift by confidence maximization
  and input transformation.
\newblock {\em arXiv preprint arXiv:2106.14999}, 2021.

\bibitem{ren2015faster}
Shaoqing Ren, Kaiming He, Ross Girshick, and Jian Sun.
\newblock Faster r-cnn: Towards real-time object detection with region proposal
  networks.
\newblock In {\em Advances in neural information processing systems}, pages
  91--99, 2015.

\bibitem{synthia}
German Ros, Laura Sellart, Joanna Materzynska, David Vazquez, and Antonio~M
  Lopez.
\newblock The synthia dataset: A large collection of synthetic images for
  semantic segmentation of urban scenes.
\newblock In {\em Proceedings of the IEEE conference on computer vision and
  pattern recognition}, pages 3234--3243, 2016.

\bibitem{roth2021simultaneous}
Karsten Roth, Timo Milbich, Bjorn Ommer, Joseph~Paul Cohen, and Marzyeh
  Ghassemi.
\newblock Simultaneous similarity-based self-distillation for deep metric
  learning.
\newblock In {\em International Conference on Machine Learning}, pages
  9095--9106. PMLR, 2021.

\bibitem{saenko2010adapting}
Kate Saenko, Brian Kulis, Mario Fritz, and Trevor Darrell.
\newblock Adapting visual category models to new domains.
\newblock In {\em European conference on computer vision}, pages 213--226.
  Springer, 2010.

\bibitem{sup_da_1}
Kuniaki Saito, Donghyun Kim, Stan Sclaroff, Trevor Darrell, and Kate Saenko.
\newblock Semi-supervised domain adaptation via minimax entropy.
\newblock In {\em Proceedings of the IEEE/CVF International Conference on
  Computer Vision}, pages 8050--8058, 2019.

\bibitem{saito2019strong}
Kuniaki Saito, Yoshitaka Ushiku, Tatsuya Harada, and Kate Saenko.
\newblock Strong-weak distribution alignment for adaptive object detection.
\newblock In {\em Proceedings of the IEEE/CVF Conference on Computer Vision and
  Pattern Recognition}, pages 6956--6965, 2019.

\bibitem{foggy}
Christos Sakaridis, Dengxin Dai, and Luc Van~Gool.
\newblock Semantic foggy scene understanding with synthetic data.
\newblock {\em International Journal of Computer Vision}, 126(9):973--992,
  2018.

\bibitem{sankaranarayanan2018generate}
Swami Sankaranarayanan, Yogesh Balaji, Carlos~D Castillo, and Rama Chellappa.
\newblock Generate to adapt: Aligning domains using generative adversarial
  networks.
\newblock In {\em Proceedings of the IEEE Conference on Computer Vision and
  Pattern Recognition}, pages 8503--8512, 2018.

\bibitem{shen2018wasserstein}
Jian Shen, Yanru Qu, Weinan Zhang, and Yong Yu.
\newblock Wasserstein distance guided representation learning for domain
  adaptation.
\newblock In {\em Thirty-Second AAAI Conference on Artificial Intelligence},
  2018.

\bibitem{shen2019scl}
Zhiqiang Shen, Harsh Maheshwari, Weichen Yao, and Marios Savvides.
\newblock Scl: Towards accurate domain adaptive object detection via gradient
  detach based stacked complementary losses.
\newblock {\em arXiv preprint arXiv:1911.02559}, 2019.

\bibitem{vgg}
Karen Simonyan and Andrew Zisserman.
\newblock Very deep convolutional networks for large-scale image recognition.
\newblock {\em arXiv preprint arXiv:1409.1556}, 2014.

\bibitem{crvae}
Samarth Sinha and Adji~B Dieng.
\newblock Consistency regularization for variational auto-encoders.
\newblock {\em arXiv preprint arXiv:2105.14859}, 2021.

\bibitem{uniform_priors}
Samarth Sinha, Karsten Roth, Anirudh Goyal, Marzyeh Ghassemi, Hugo Larochelle,
  and Animesh Garg.
\newblock Uniform priors for data-efficient transfer.
\newblock {\em arXiv preprint arXiv:2006.16524}, 2020.

\bibitem{fixmatch}
Kihyuk Sohn, David Berthelot, Chun-Liang Li, Zizhao Zhang, Nicholas Carlini,
  Ekin~D Cubuk, Alex Kurakin, Han Zhang, and Colin Raffel.
\newblock Fixmatch: Simplifying semi-supervised learning with consistency and
  confidence.
\newblock {\em arXiv preprint arXiv:2001.07685}, 2020.

\bibitem{ttt}
Yu Sun, Xiaolong Wang, Zhuang Liu, John Miller, Alexei Efros, and Moritz Hardt.
\newblock Test-time training with self-supervision for generalization under
  distribution shifts.
\newblock In {\em International Conference on Machine Learning}, pages
  9229--9248. PMLR, 2020.

\bibitem{tang2021humble}
Yihe Tang, Weifeng Chen, Yijun Luo, and Yuting Zhang.
\newblock Humble teachers teach better students for semi-supervised object
  detection.
\newblock In {\em Proceedings of the IEEE/CVF Conference on Computer Vision and
  Pattern Recognition}, pages 3132--3141, 2021.

\bibitem{tarvainen2017mean}
Antti Tarvainen and Harri Valpola.
\newblock Mean teachers are better role models: Weight-averaged consistency
  targets improve semi-supervised deep learning results.
\newblock {\em arXiv preprint arXiv:1703.01780}, 2017.

\bibitem{adda}
Eric Tzeng, Judy Hoffman, Kate Saenko, and Trevor Darrell.
\newblock Adversarial discriminative domain adaptation.
\newblock In {\em Proceedings of the IEEE conference on computer vision and
  pattern recognition}, pages 7167--7176, 2017.

\bibitem{tzeng2014deep}
Eric Tzeng, Judy Hoffman, Ning Zhang, Kate Saenko, and Trevor Darrell.
\newblock Deep domain confusion: Maximizing for domain invariance.
\newblock {\em arXiv preprint arXiv:1412.3474}, 2014.

\bibitem{wang2021target}
Dequan Wang, Shaoteng Liu, Sayna Ebrahimi, Evan Shelhamer, and Trevor Darrell.
\newblock On-target adaptation.
\newblock {\em arXiv preprint arXiv:2109.01087}, 2021.

\bibitem{tent}
Dequan Wang, Evan Shelhamer, Shaoteng Liu, Bruno Olshausen, and Trevor Darrell.
\newblock Tent: Fully test-time adaptation by entropy minimization.
\newblock {\em arXiv preprint arXiv:2006.10726}, 2020.

\bibitem{xie2020self}
Qizhe Xie, Minh-Thang Luong, Eduard Hovy, and Quoc~V Le.
\newblock Self-training with noisy student improves imagenet classification.
\newblock In {\em Proceedings of the IEEE/CVF Conference on Computer Vision and
  Pattern Recognition}, pages 10687--10698, 2020.

\bibitem{xu2020exploring}
Chang-Dong Xu, Xing-Ran Zhao, Xin Jin, and Xiu-Shen Wei.
\newblock Exploring categorical regularization for domain adaptive object
  detection.
\newblock In {\em Proceedings of the IEEE/CVF Conference on Computer Vision and
  Pattern Recognition}, pages 11724--11733, 2020.

\bibitem{xu2020cross}
Minghao Xu, Hang Wang, Bingbing Ni, Qi Tian, and Wenjun Zhang.
\newblock Cross-domain detection via graph-induced prototype alignment.
\newblock In {\em Proceedings of the IEEE/CVF Conference on Computer Vision and
  Pattern Recognition}, pages 12355--12364, 2020.

\bibitem{xu2021end}
Mengde Xu, Zheng Zhang, Han Hu, Jianfeng Wang, Lijuan Wang, Fangyun Wei, Xiang
  Bai, and Zicheng Liu.
\newblock End-to-end semi-supervised object detection with soft teacher.
\newblock {\em arXiv preprint arXiv:2106.09018}, 2021.

\bibitem{sup_da_2}
Ting Yao, Yingwei Pan, Chong-Wah Ngo, Houqiang Li, and Tao Mei.
\newblock Semi-supervised domain adaptation with subspace learning for visual
  recognition.
\newblock In {\em Proceedings of the IEEE conference on Computer Vision and
  Pattern Recognition}, pages 2142--2150, 2015.

\bibitem{ye2021few}
Han-Jia Ye, Lu Ming, De-Chuan Zhan, and Wei-Lun Chao.
\newblock Few-shot learning with a strong teacher.
\newblock {\em arXiv preprint arXiv:2107.00197}, 2021.

\bibitem{drn}
Fisher Yu, Vladlen Koltun, and Thomas Funkhouser.
\newblock Dilated residual networks.
\newblock In {\em Proceedings of the IEEE conference on computer vision and
  pattern recognition}, pages 472--480, 2017.

\bibitem{bdd}
Fisher Yu, Wenqi Xian, Yingying Chen, Fangchen Liu, Mike Liao, Vashisht
  Madhavan, and Trevor Darrell.
\newblock Bdd100k: A diverse driving video database with scalable annotation
  tooling.
\newblock {\em arXiv preprint arXiv:1805.04687}, 2(5):6, 2018.

\bibitem{zhang2019synthetic}
Hui Zhang, Yonglin Tian, Kunfeng Wang, Haibo He, and Fei-Yue Wang.
\newblock Synthetic-to-real domain adaptation for object instance segmentation.
\newblock In {\em 2019 International Joint Conference on Neural Networks
  (IJCNN)}, pages 1--7. IEEE, 2019.

\bibitem{zhao2019learning}
Han Zhao, Remi~Tachet Des~Combes, Kun Zhang, and Geoffrey Gordon.
\newblock On learning invariant representations for domain adaptation.
\newblock In {\em International Conference on Machine Learning}, pages
  7523--7532. PMLR, 2019.

\bibitem{zhou2020learning}
Kaiyang Zhou, Yongxin Yang, Timothy Hospedales, and Tao Xiang.
\newblock Learning to generate novel domains for domain generalization.
\newblock In {\em European Conference on Computer Vision}, pages 561--578.
  Springer, 2020.

\bibitem{deformable_detr}
Xizhou Zhu, Weijie Su, Lewei Lu, Bin Li, Xiaogang Wang, and Jifeng Dai.
\newblock Deformable detr: Deformable transformers for end-to-end object
  detection.
\newblock {\em arXiv preprint arXiv:2010.04159}, 2020.

\end{thebibliography}
}
\clearpage

\newpage
\newpage

\appendix
\section{Task Description}
\label{app:tasks}

\begin{table}[h!]
    \centering
    \begin{tabular}{l|cc}
        \toprule
        \textbf{Task Description} & \textbf{$\#$ Train} & \textbf{$\#$ Test} \\
        & \textbf{Images} & \textbf{Images} \\
        \midrule
        \multicolumn{3}{l}{\textbf{Object Detection}}\\
        Sim10k $\rightarrow$ Cityscapes & 10000 & 25000 \\
        Cityscapes $\rightarrow$ Foggy & 20000 & 5000 \\
        Cityscapes $\rightarrow$ KITTI & 25000 & 7481 \\
        Cityscapes $\rightarrow$ BDD100k & 25000 & 10000 \\
        MS-COCO 4 $\rightarrow$ 1 & 94400 & 23600 \\
        MS-COCO 9 $\rightarrow$ 1 & 106200 & 11800 \\
        \midrule
        \multicolumn{3}{l}{\textbf{Image Segmentation}} \\
        GTA $\rightarrow$ Cityscapes & 25000 & 2975 \\
        Synthia $\rightarrow$ Cityscapes & 9400 & 2975 \\
        \bottomrule
    \end{tabular}
    \caption{Description of the tasks, number of train images and number of test images required to train the 
    vanilla unsupervised domain adaptation algorithms, which require access to the test-images at training time
    \cite{saito2019strong,chen2020harmonizing}}
    \label{tab:task-desc}
\end{table}

The number of training and testing images for each dataset split is available in Table \ref{tab:task-desc}.
The number of testing images describes the number of unlabeled images that are available at training time
for the baselines ``Chen et al.'' \cite{chen2020harmonizing} and ``Saito et al.'' \cite{saito2019strong}
for the vanilla unsupervised object detection results, since both methods assume access to the test distribution at training time. 

\section{Further Experiments}
\label{app:further-exp}

\begin{table}[h!]
\centering
\begin{tabular}{l|c c c c}
    \toprule
    \textbf{Task} & \textbf{TTT} & \textbf{Tent} & \textbf{Ours} \\
    \midrule
        Sim10k $\rightarrow$ Cityscapes & 13.2 & 15.6 & \textbf{8.9} \\
        Cityscapes $\rightarrow$ Foggy Cityscapes  & 10.4 & 11.3 & \textbf{9.0} \\
        Cityscapes $\rightarrow$ KITTI  & 10.5 & 11.8 & \textbf{8.5} \\
        Cityscapes $\rightarrow$ BDD100k & 9.6 & 10.2 & \textbf{8.3} \\
    \bottomrule
\end{tabular}
\caption{\textbf{Detection-Expected Calibration Error (d-ECE)} of the models on the test set. Lower d-ECE is better.
All models are trained on the novel distributions with a budget ($n$) = 64.
We see that using TeST, we not only improve the accuracy performance of the models, but also are able to
improve the calibration of the models predictions.}
\label{tab:dece}
\end{table}

\paragraph{Calibration of Predictions}
One metric that is often overlooked when deploying predictive models is that of calibration: 
are the probability scores calibrated to their performance.
Similar to accuracy, we also analyse the effect of test-time adaptation on the calibration of the
resultant models. 
We use the recent Detection-Expected Calibration Error (D-ECE) metric to measure the calibration
of the predictions \cite{dece}.
To test this, we experiment with the same self-drivimg domain adaptation benchmarks we used for
the object detection using a base Faster-RCNN detector. 
The results are presented in Table \ref{tab:dece}.
Interestingly, we see that using pseudo-labels from the teacher, we are able to
outperform both TTT and Tent by being better calibrated.
By performing entropy minimization and knowledge distillation on the student, we are able to get 
better performance, and model predictions that are better calibrated.

\section{Qualitative Results}

\begin{figure*}[t!]
    \centering
    \begin{subfigure}{0.8\textwidth}
    \includegraphics[width=\textwidth]{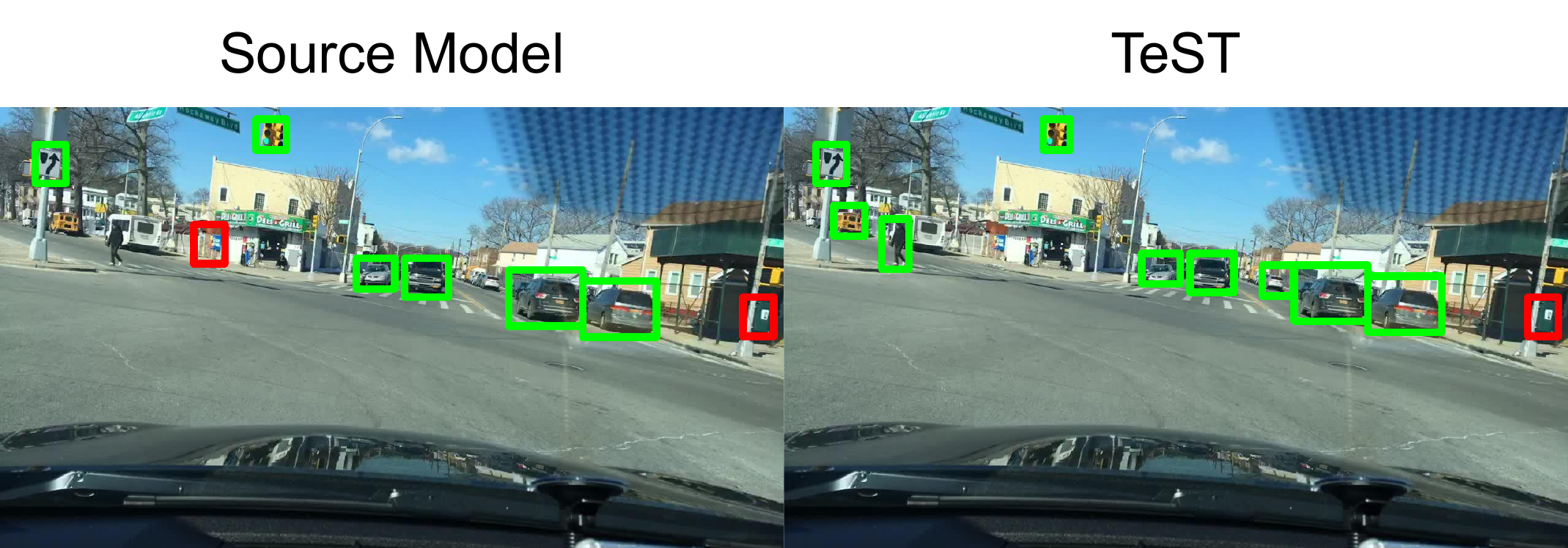}
    \end{subfigure}
    \begin{subfigure}{0.8\textwidth}
    \includegraphics[width=\textwidth]{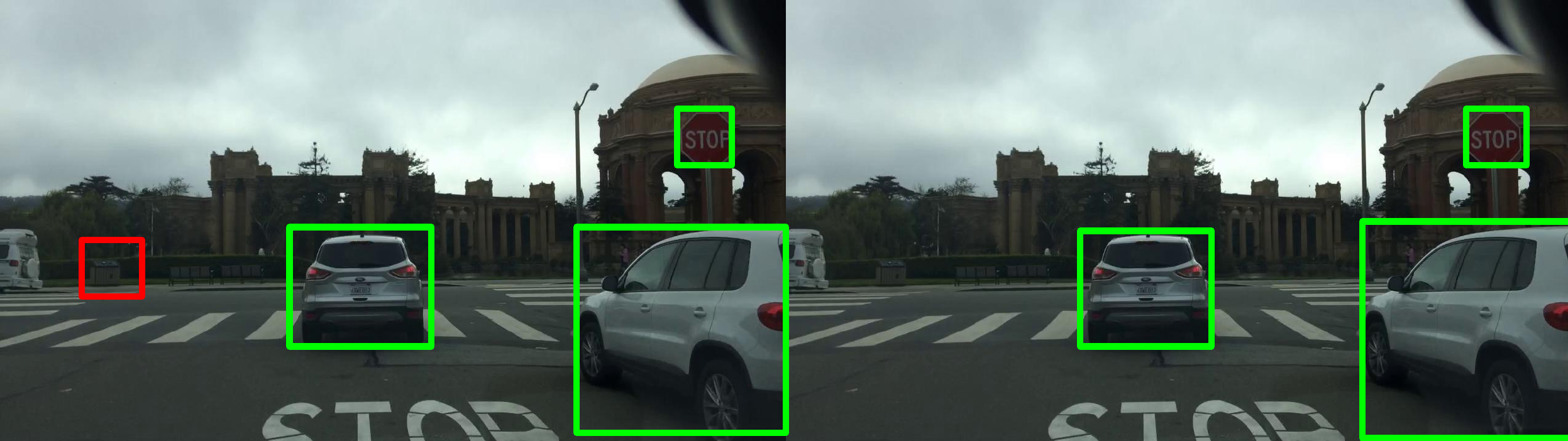}
    \end{subfigure}
    \begin{subfigure}{0.8\textwidth}
    \includegraphics[width=\textwidth]{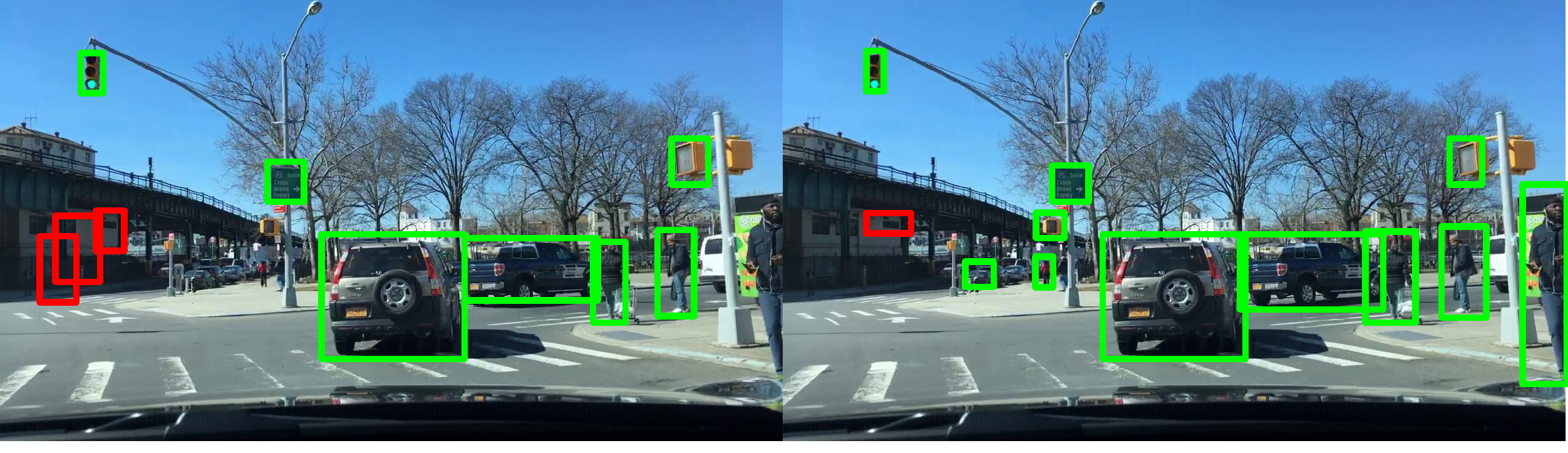}
    \end{subfigure}
    \begin{subfigure}{0.8\textwidth}
    \includegraphics[width=\textwidth]{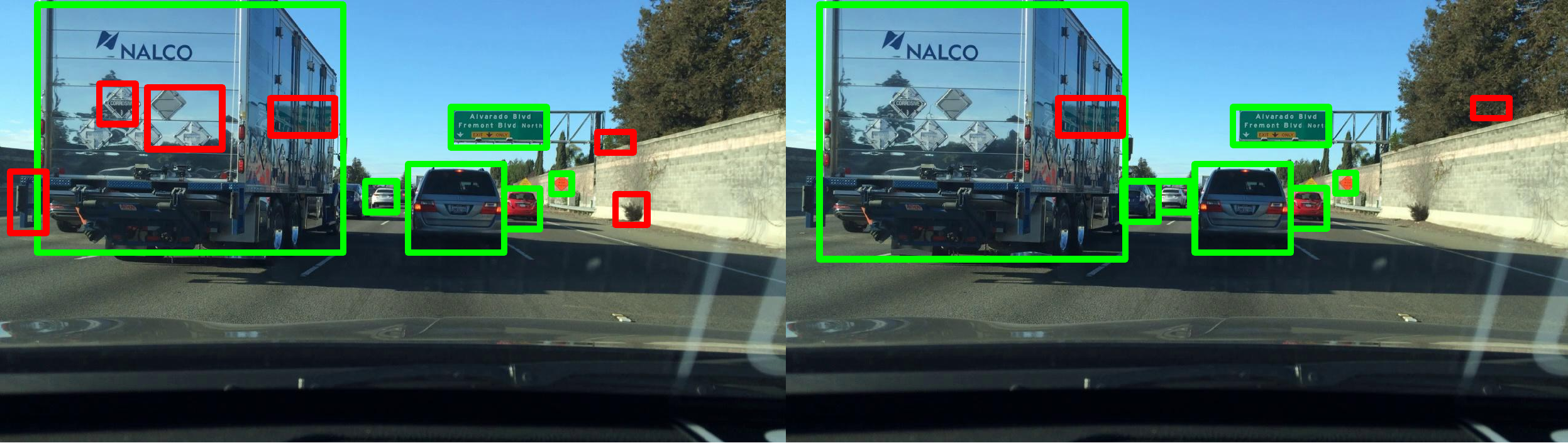}
    \end{subfigure}
    \begin{subfigure}{0.8\textwidth}
    \includegraphics[width=\textwidth]{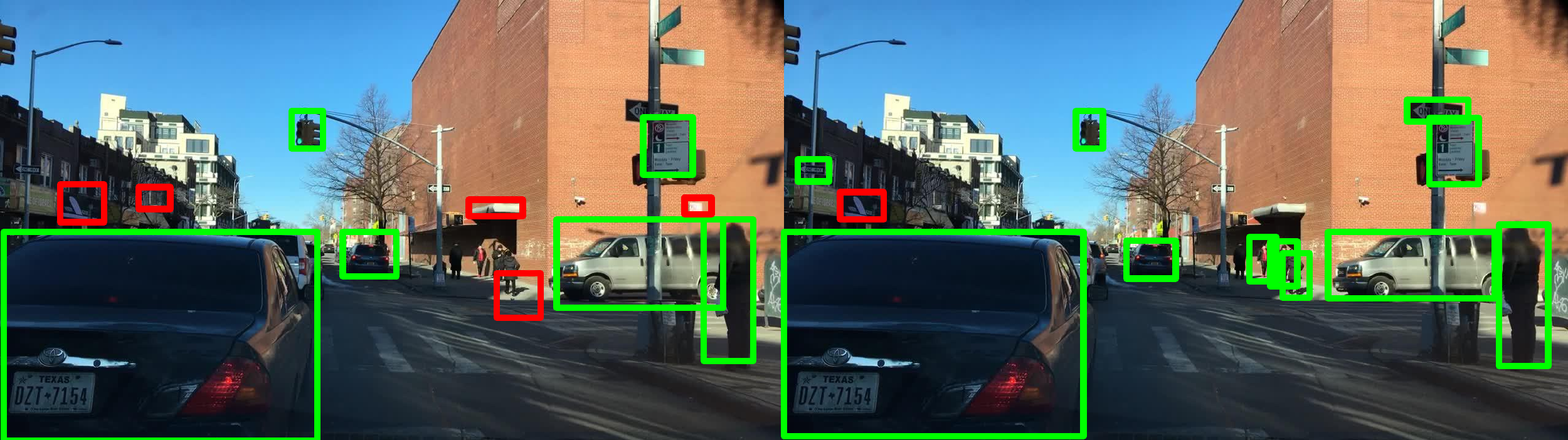}
    \end{subfigure}
    \caption{Qualitative results from BDD100k with a Faster RCNN Object Detector \cite{ren2015faster}.
    True positives are shown in {\color{green} green} rectangles and
    False positives are shown in {\color{red} red} rectangles.
    We note that: \textbf{all images are chosen at random without any cherry-picking.}
    We see that the models trained with TeST have fewer false-positives and more
    true-positives, which strongly suggests that TeST is able to 
    improve the final object detector on the novel dataset.}
    \label{fig:qual-rcnn-bdd}
\end{figure*}

\begin{figure*}[t!]
    \centering
    \begin{subfigure}{0.64\textwidth}
    \includegraphics[width=\textwidth]{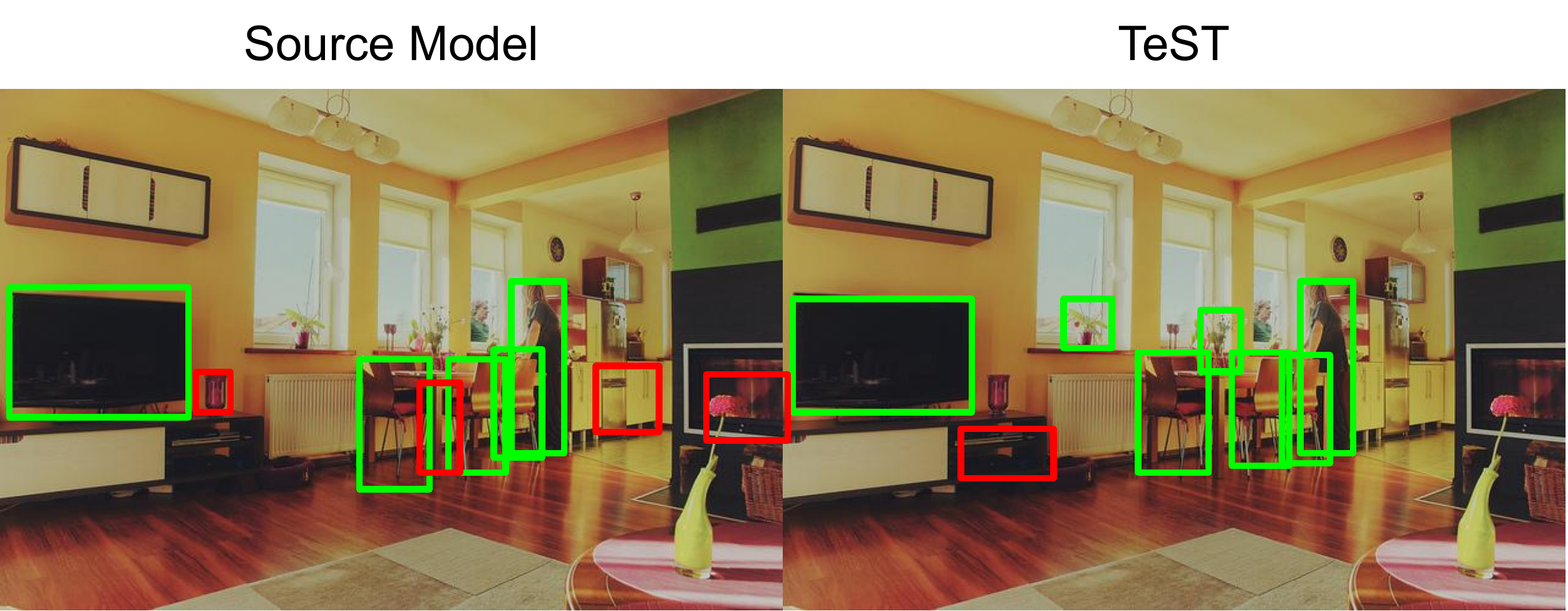}
    \end{subfigure}
    \begin{subfigure}{0.64\textwidth}
    \includegraphics[width=\textwidth]{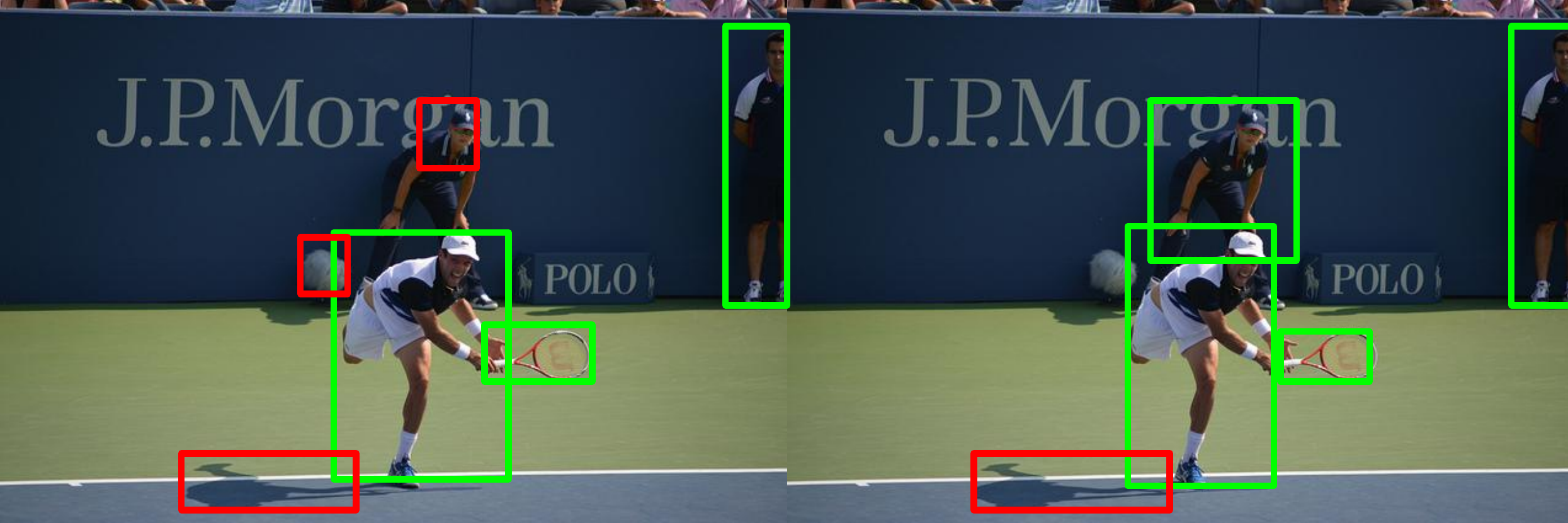}
    \end{subfigure}
    \begin{subfigure}{0.64\textwidth}
    \includegraphics[width=\textwidth]{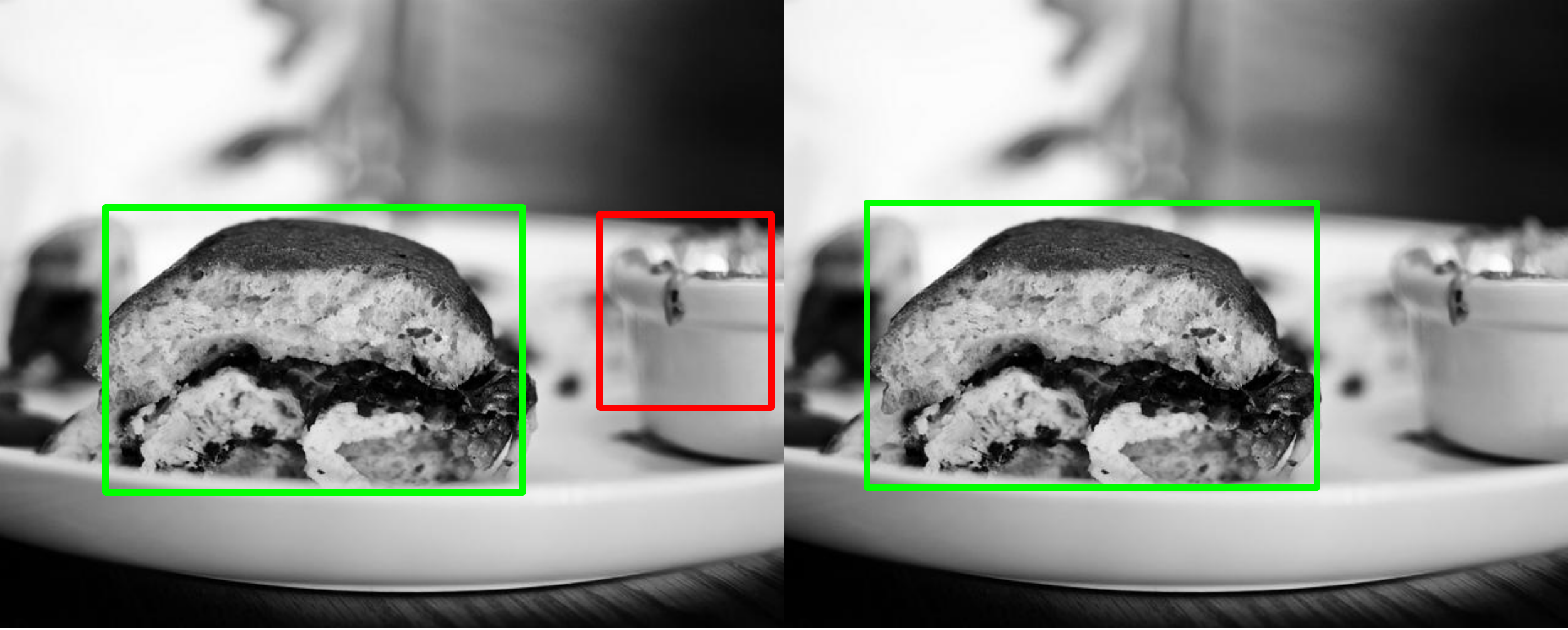}
    \end{subfigure}
    \begin{subfigure}{0.64\textwidth}
    \includegraphics[width=\textwidth]{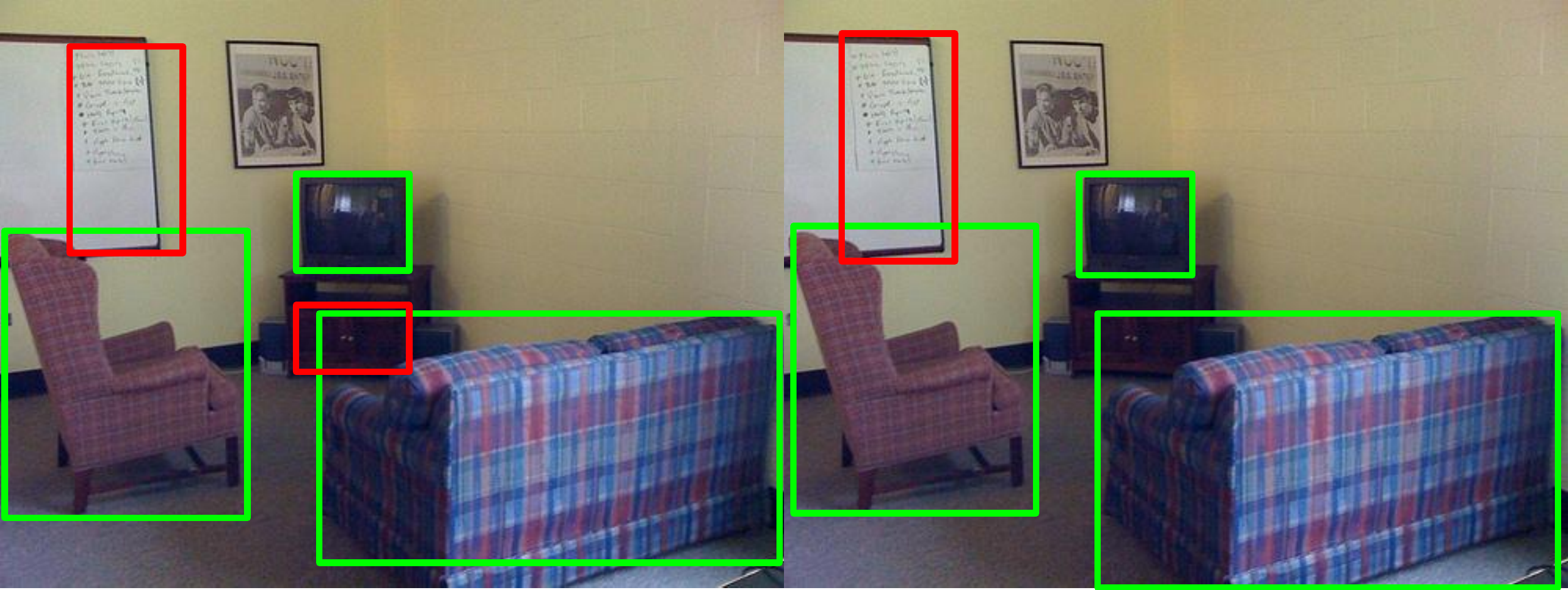}
    \end{subfigure}
    \begin{subfigure}{0.64\textwidth}
    \includegraphics[width=\textwidth]{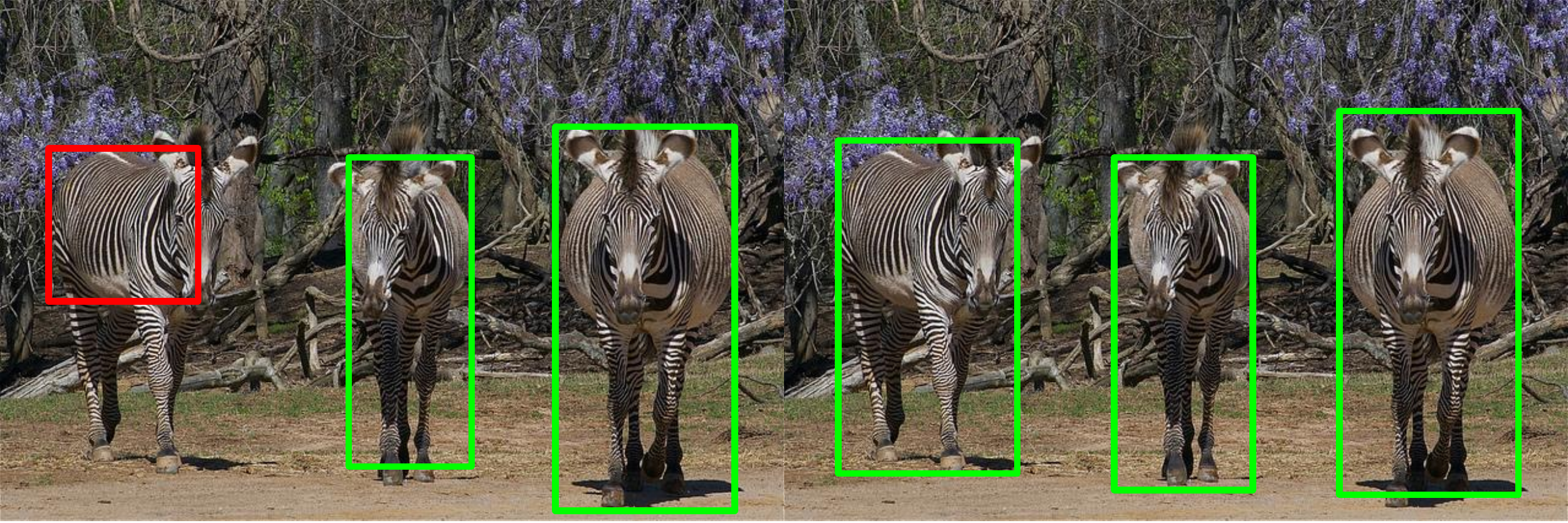}
    \end{subfigure}
    \caption{Qualitative results from MS-COCO with a Faster RCNN Object Detector
    \cite{ren2015faster}.
    True positives are shown in {\color{green} green} rectangles and
    False positives are shown in {\color{red} red} rectangles.
    We note that: \textbf{all images are chosen at random without any cherry-picking.}
    We see that the models trained with TeST have fewer false-positives and more
    true-positives, which strongly suggests that TeST is able to 
    improve the final object detector on the novel dataset.}
    \label{fig:qual-rcnn-coco}
\end{figure*}

\begin{figure*}[t!]
    \centering
    \begin{subfigure}{0.8\textwidth}
    \includegraphics[width=\textwidth]{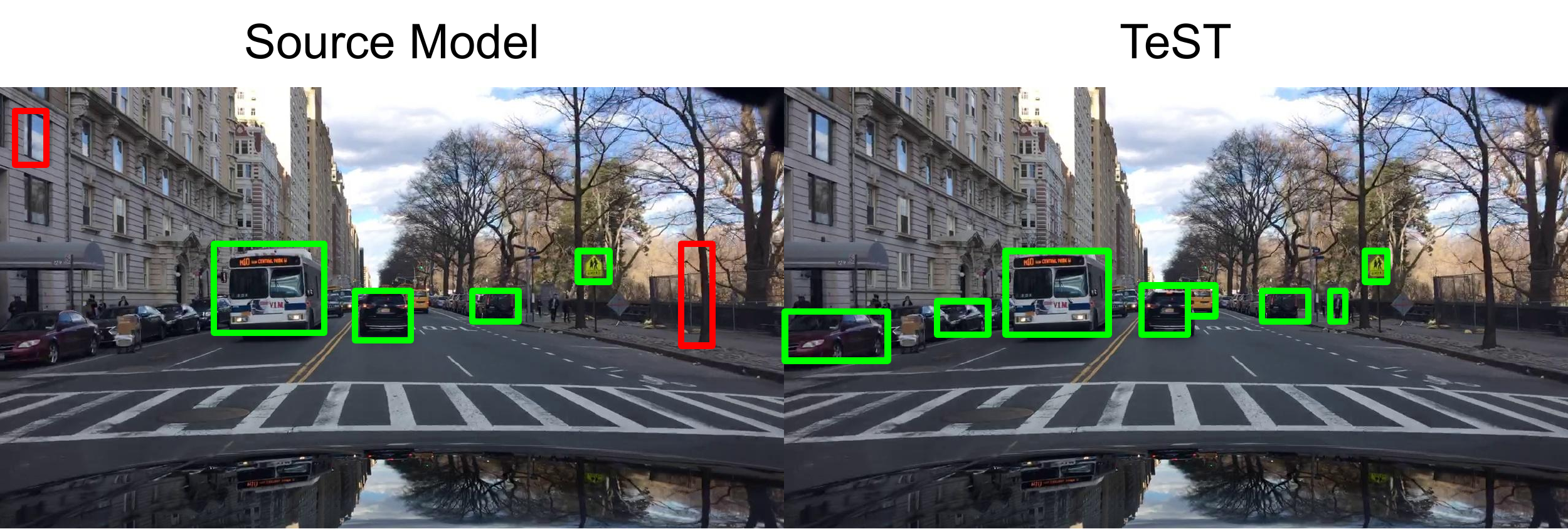}
    \end{subfigure}
    \begin{subfigure}{0.8\textwidth}
    \includegraphics[width=\textwidth]{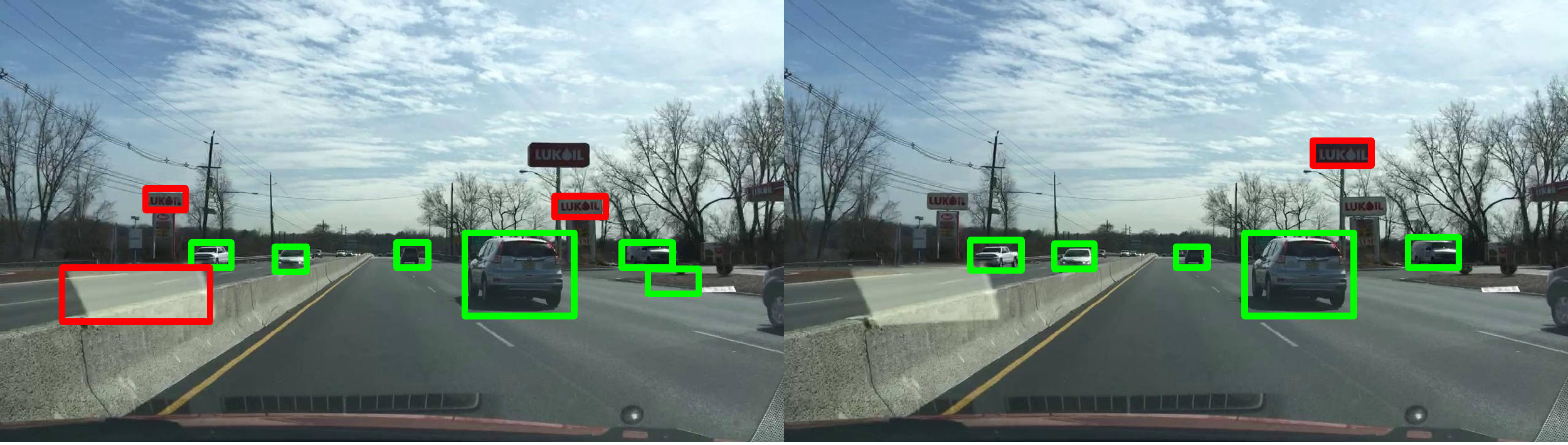}
    \end{subfigure}
    \begin{subfigure}{0.8\textwidth}
    \includegraphics[width=\textwidth]{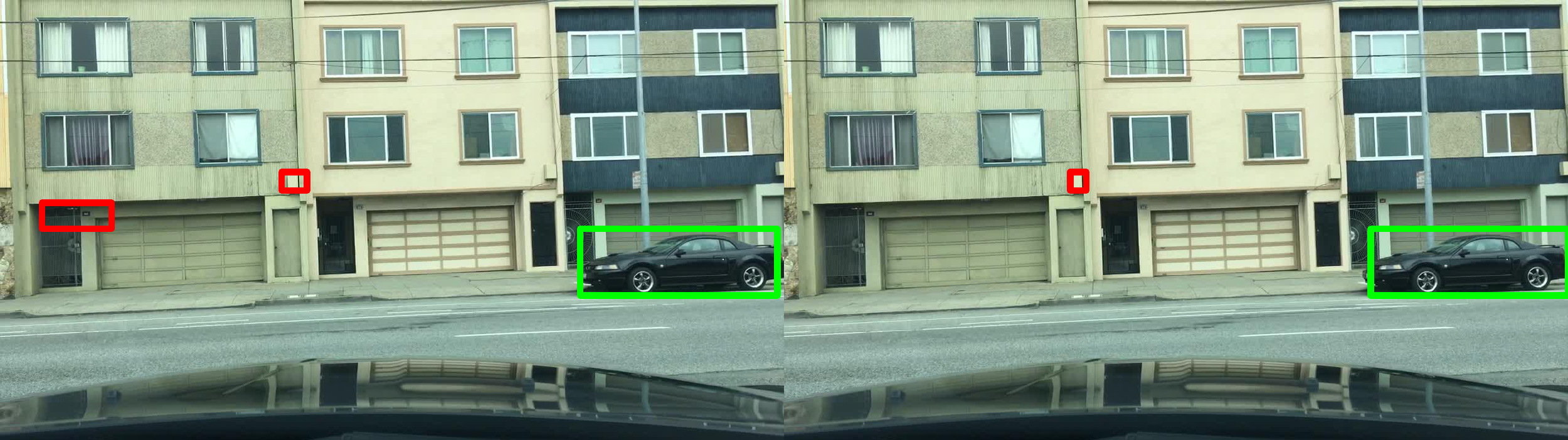}
    \end{subfigure}
    \begin{subfigure}{0.8\textwidth}
    \includegraphics[width=\textwidth]{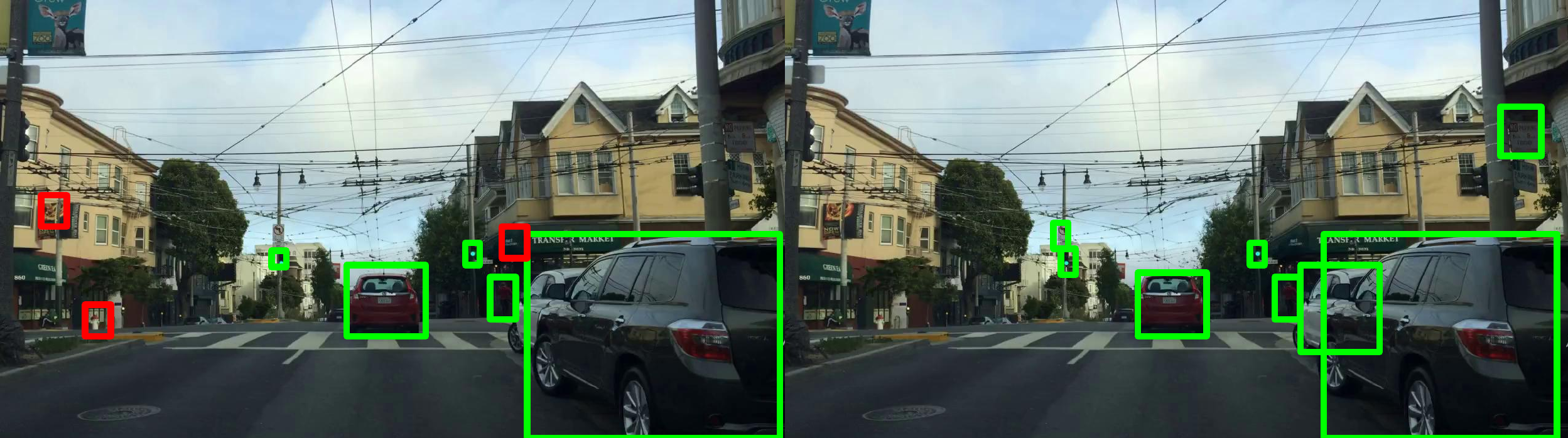}
    \end{subfigure}
    \begin{subfigure}{0.8\textwidth}
    \includegraphics[width=\textwidth]{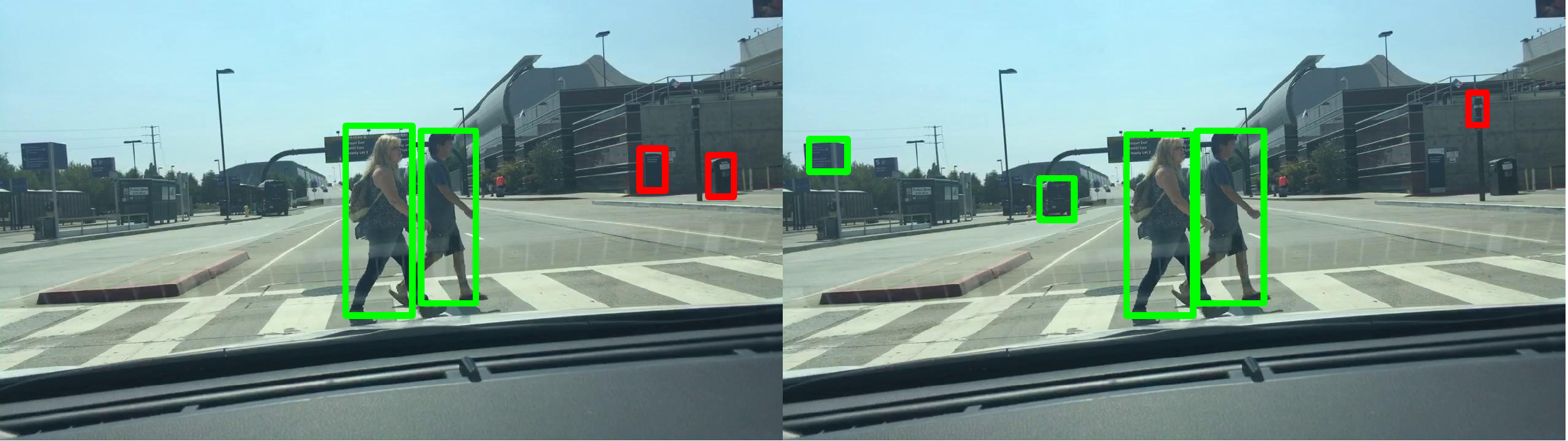}
    \end{subfigure}
    \caption{Qualitative results from BDD100k with a Deformable DeTR Object Detector 
    \cite{deformable_detr}.
    True positives are shown in {\color{green} green} rectangles and
    False positives are shown in {\color{red} red} rectangles.
    We note that: \textbf{all images are chosen at random without any cherry-picking.}
    We see that the models trained with TeST have fewer false-positives and more
    true-positives, which strongly suggests that TeST is able to 
    improve the final object detector on the novel dataset.
    We see that the models trained with TeST have fewer false-positives and more
    true-positives, which strongly suggests that TeST is able to 
    improve the final object detector on the novel dataset.}
    \label{fig:qual-detr-bdd}
\end{figure*}

\begin{figure*}[t!]
    \centering
    \begin{subfigure}{0.65\textwidth}
    \includegraphics[width=\textwidth]{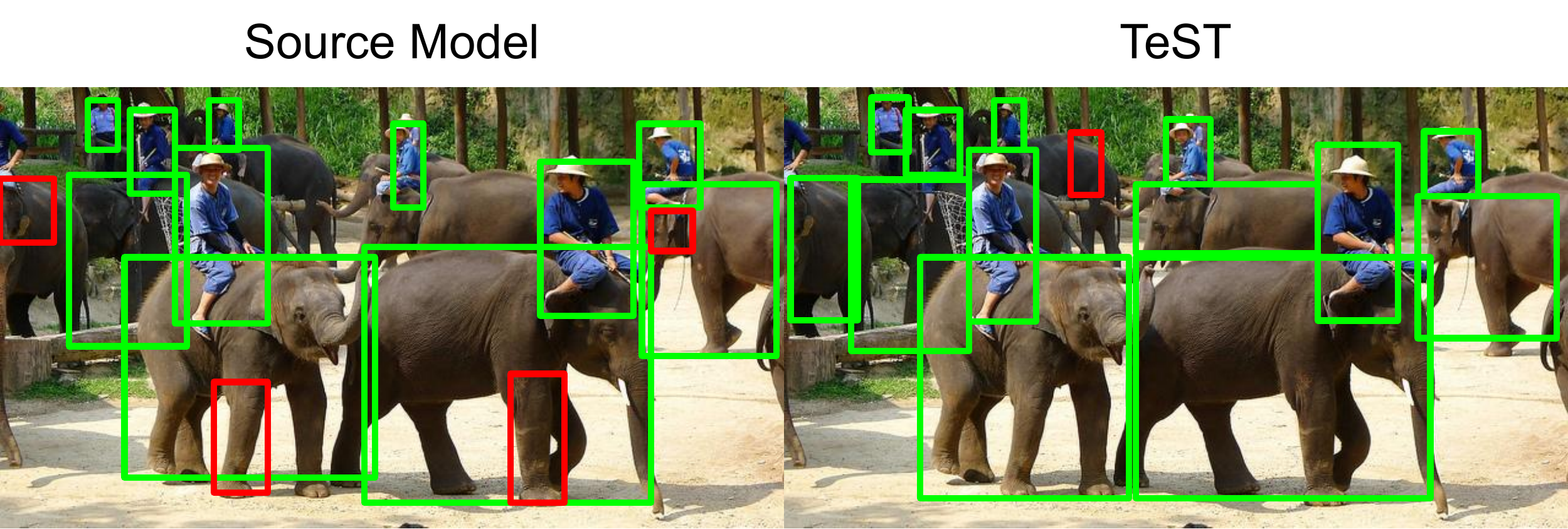}
    \end{subfigure}
    \begin{subfigure}{0.65\textwidth}
    \includegraphics[width=\textwidth]{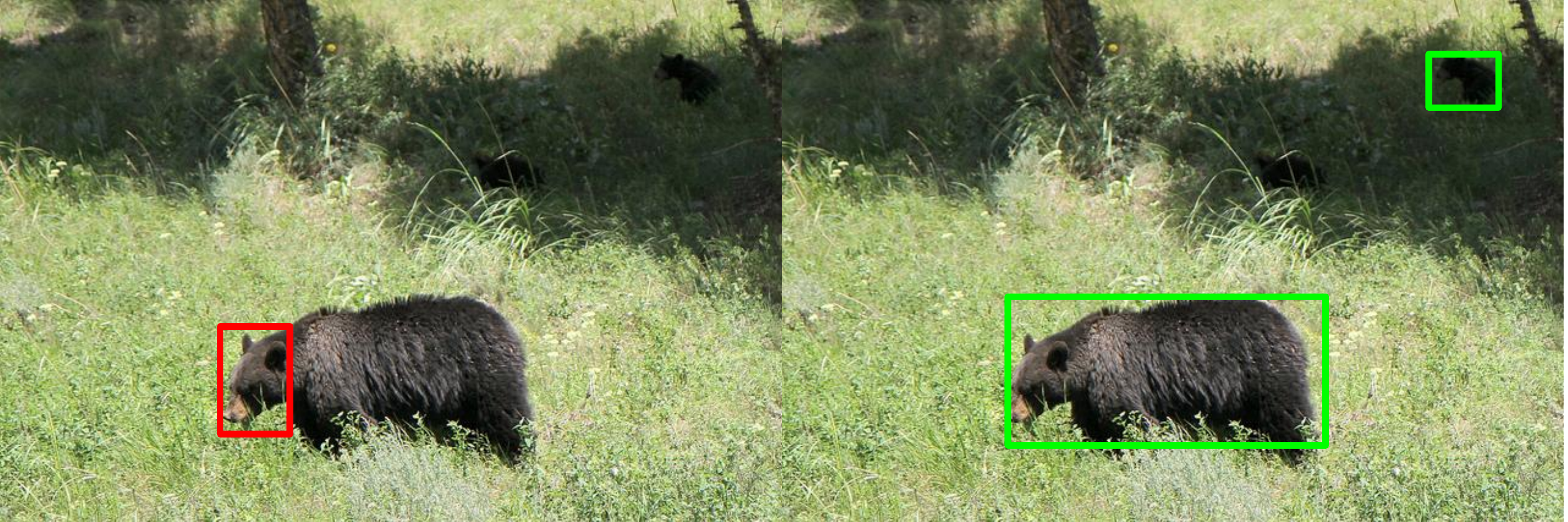}
    \end{subfigure}
    \begin{subfigure}{0.65\textwidth}
    \includegraphics[width=\textwidth]{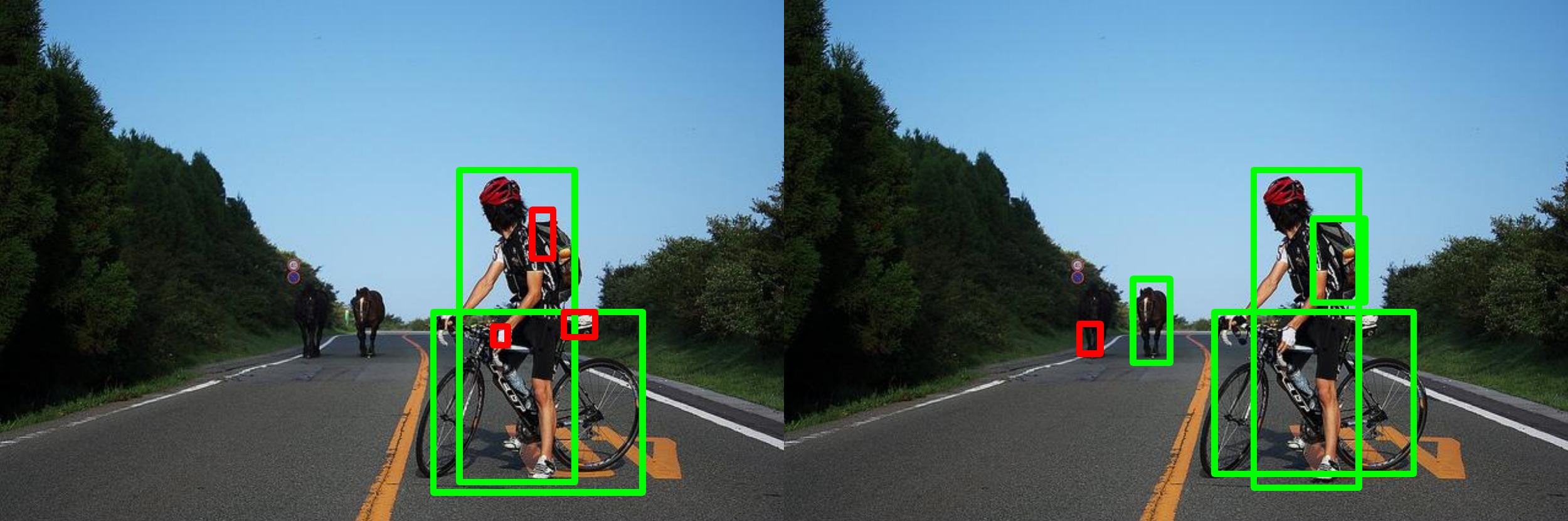}
    \end{subfigure}
    \begin{subfigure}{0.65\textwidth}
    \includegraphics[width=\textwidth]{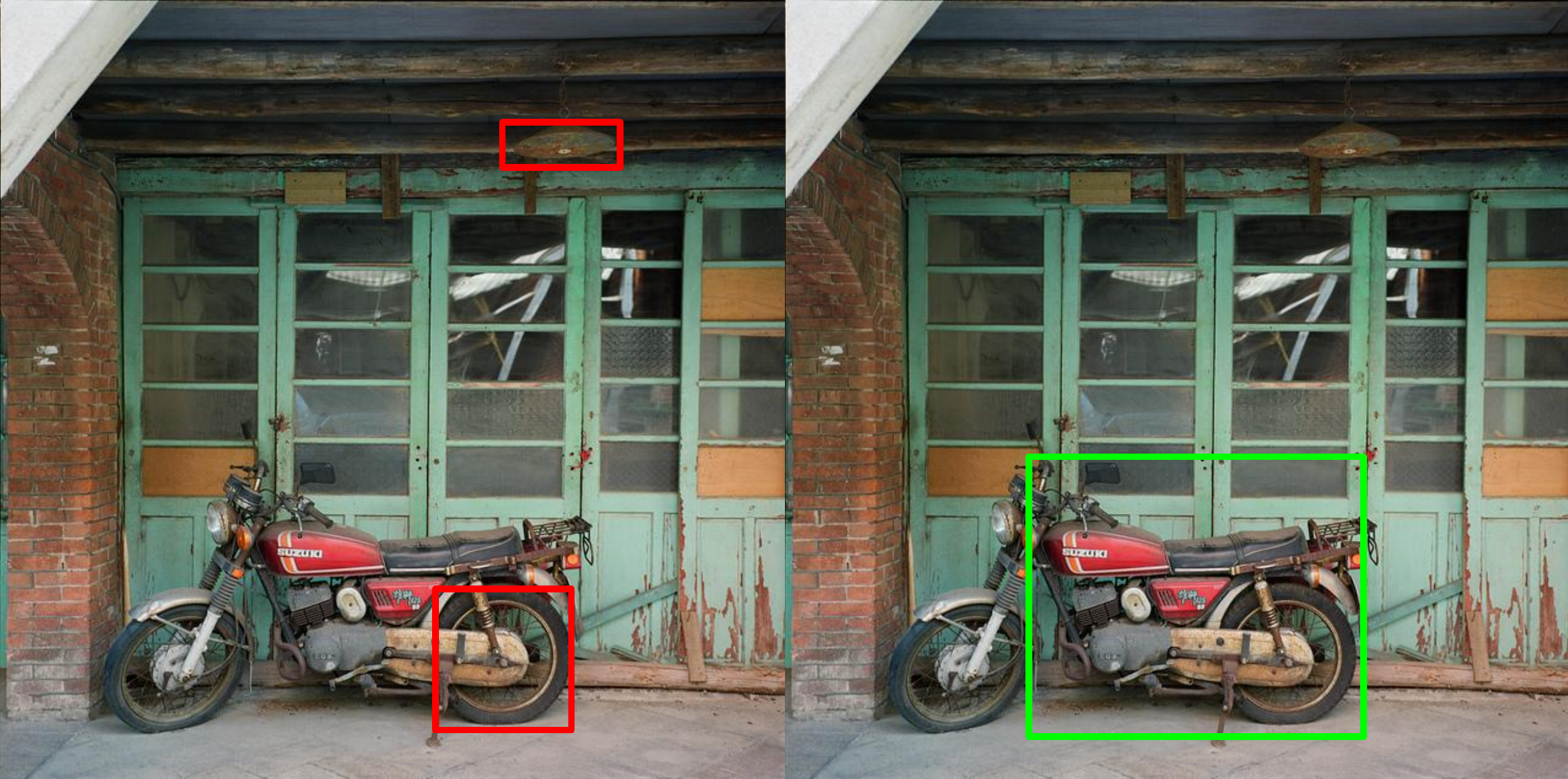}
    \end{subfigure}
    \begin{subfigure}{0.65\textwidth}
    \includegraphics[width=\textwidth]{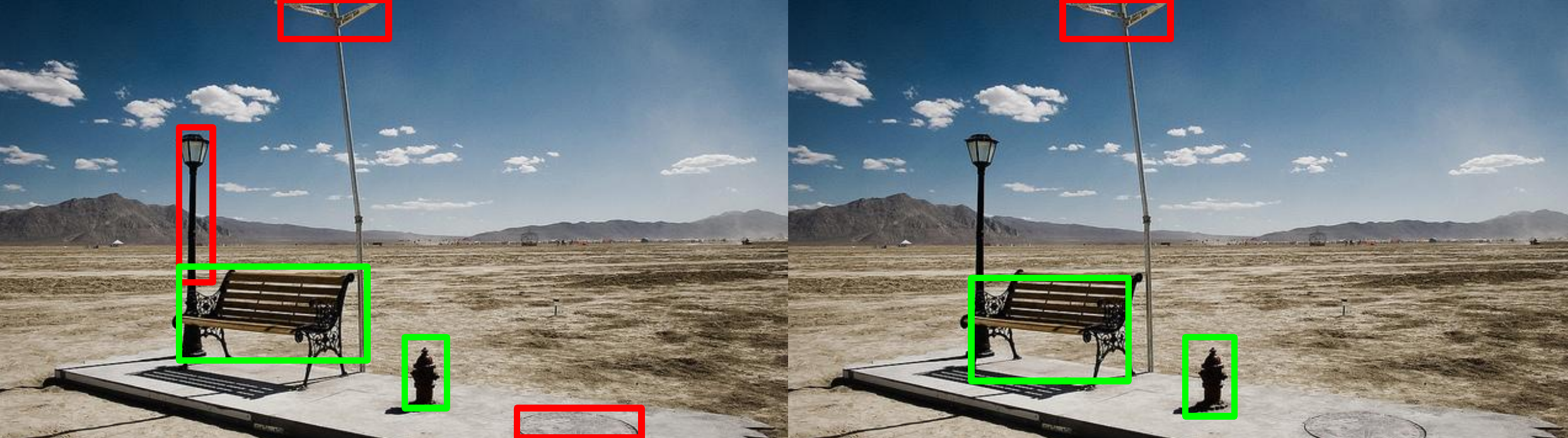}
    \end{subfigure}
    \caption{Qualitative results from MS-COCO with a Deformable DeTR Object Detector 
    \cite{deformable_detr}.
    True positives are shown in {\color{green} green} rectangles and
    False positives are shown in {\color{red} red} rectangles.
    We note that: \textbf{all images are chosen at random without any cherry-picking.}
    We see that the models trained with TeST have fewer false-positives and more
    true-positives, which strongly suggests that TeST is able to 
    improve the final object detector on the novel dataset.}
    \label{fig:qual-detr-coco}
\end{figure*}

Along with the quantitative results, we perform a thorough qualitative evaluation of TeST.
We first further investigation the effect of adding TeST by investigating the true-postives and
false-positive outputs from the object detectors.
Figures \ref{fig:qual-rcnn-bdd} and \ref{fig:qual-rcnn-coco} show results for a base Faster 
RCNN object detector \cite{ren2015faster} on the BDD100k \cite{bdd} and MS-COCO \cite{coco} datasets, 
respectively before and after TeST.
Figures \ref{fig:qual-detr-bdd} and \ref{fig:qual-detr-coco} show results for a base 
Deformable Detection Transformer object detector \cite{deformable_detr} on the BDD100k \cite{bdd}
and MS-COCO \cite{coco} datasets, respectively, before and after TeST.
All the images are randomly sampled, and \textbf{we do not perform any cherry-picking to get better qualitative
results}.
We see that in each of the examples, by adding TeST, we are able to increase the number 
of true positives, while decreasing the number of false positives, both of which are important
qualities of a good object detector. 
By performing qualitative examples on a driving dataset (BDD-100k) and a \textit{common objects}
datset (MS-COCO), we are able to evaluate the qualitative improvements on both fronts.

\begin{figure*}
    \centering
    \includegraphics[width=\textwidth]{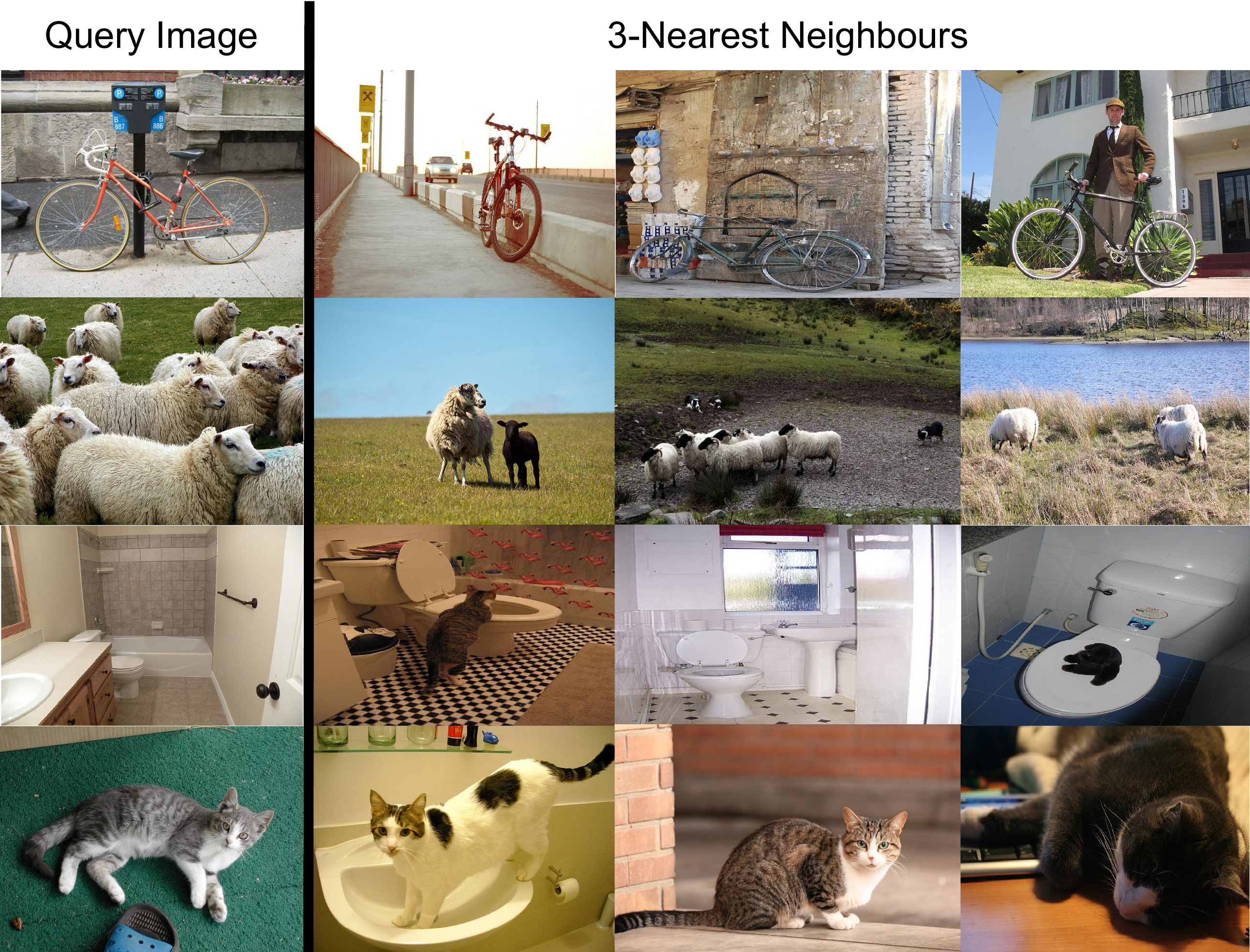}
    \caption{3-Nearest Neighbours in the embedding space
    for the feature extractor of a
    Faster-RCNN detector \cite{ren2015faster}, after it has been trained
    using TeST on the COCO dataset. 
    We see that the model is able to learn semantically 
    meaningful representations and the nearest neighbours to the query image
    are semantically similar, thereby showing that TeST is able to perform
    meaningful reprensentation learning.}
    \label{fig:knn-rcnn}
\end{figure*}

\begin{figure*}
    \centering
    \includegraphics[width=\textwidth]{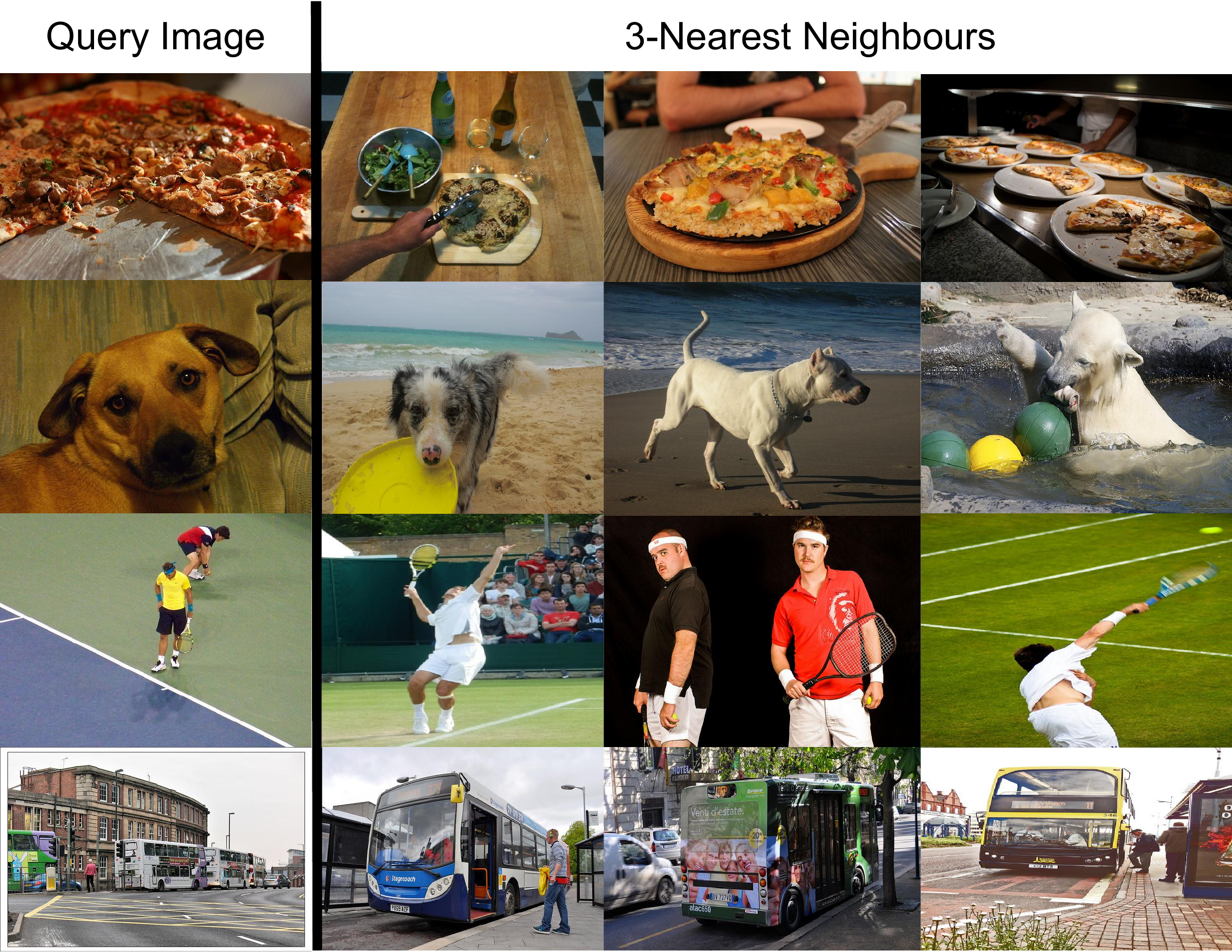}
    \caption{3-Nearest Neighbours in the embedding space
    for the feature extractor of a
    Deformable DeTR detector \cite{deformable_detr}, after it has been trained
    using TeST on the COCO dataset. 
    We see that the model is able to learn semantically 
    meaningful representations and the nearest neighbours to the query image
    are semantically similar, thereby showing that TeST is able to perform
    meaningful reprensentation learning.}
    \label{fig:knn-detr}
\end{figure*}

Furthermore, we also investigate the representations learned by TeST by looking at the 3-nearest
neighbours in the test set for a given \textit{query image}. 
Figure \ref{fig:knn-rcnn} and \ref{fig:knn-detr} show the 3-nearest neighbour results for the
MS-COCO dataset for a Faster RCNN \cite{ren2015faster} and a Deformable DeTR \cite{deformable_detr},
respectively. 
The query and neighbour images are from the test-set.
We see that using TeST, the model consistently learns semantically relevant features, as the 
nearest neighbours are from the same class as the objects in the query image.

\end{document}